\documentclass[11pt, letterpaper, logo, onecolumn, copyright]{main}

\usepackage[authoryear, sort&compress, round]{natbib}
\usepackage[inkscapeformat=png]{svg}
\usepackage[most, breakable, skins]{tcolorbox}
\tcbuselibrary{listings,skins,breakable}

\usepackage{lipsum}
\usepackage{tabularx}
\usepackage{afterpage}
\usepackage{booktabs}
\usepackage{subcaption}
\usepackage{makecell}
\usepackage{multirow}
\usepackage{bm}
\usepackage{multicol}
\usepackage{array}
\usepackage{float}
\usepackage{listings}
\IfFileExists{listings-rust.sty}{\usepackage{listings-rust}}{}
\usepackage{fontawesome5}
\usepackage{hyperref}
\usepackage{amssymb,graphicx}
\usepackage[dvipsnames]{xcolor}
\usepackage{cleveref}
\usepackage{longtable}
\usepackage{pdflscape}
\usepackage{adjustbox}
\usepackage{nicematrix}
\usepackage{CJKutf8}
\usepackage{ragged2e}
\usepackage{colortbl}
\usepackage{enumitem}
\usepackage{algorithm}
\usepackage{algorithmic}
\usepackage{pifont}
\usepackage[htt]{hyphenat}
\usepackage{amsmath}
\usepackage{amsthm}
\usepackage{mathtools}
\usepackage{mathrsfs}
\usepackage{circledsteps}
\usepackage{diagbox}
\usepackage{bookmark}
\usepackage{wrapfig}
\usepackage{xspace}
\usepackage{tikz}
\usepackage[normalem]{ulem}
\usepackage{flafter}
\usepackage{placeins}
\usepackage{pgfplots}
\usepackage{pgfplotstable}

\usepackage{docmute}
\usepackage{environ}

\makeatletter
\let\PaperIncludeGraphics\includegraphics
\renewcommand{\includegraphics}[2][]{%
  \IfFileExists{#2}{%
    \PaperIncludeGraphics[#1]{#2}%
  }{%
    \begingroup
      \setkeys{Gin}{#1}%
      \dimen@=\Gin@req@width
      \ifdim\dimen@>\linewidth
        \dimen@=\linewidth
      \fi
      \setlength{\fboxsep}{0pt}%
      \fbox{%
        \parbox[c][.28\dimen@][c]{\dimexpr\dimen@-2\fboxrule\relax}{%
          \centering\sloppy\footnotesize\ttfamily\detokenize{#2}%
        }%
      }%
    \endgroup
  }%
}
\makeatother

\pretocmd{\section}{}{}{}

\pgfplotsset{compat=1.18}
\lstdefinelanguage{promptlang}{
  morekeywords={},
  sensitive=true,
  literate=
    {—}{{---}}1
    {–}{{--}}1
    {~}{{\textasciitilde}}1
    {…}{{\ldots}}1
    {→}{{$\rightarrow$}}1
    {✓}{{\checkmark}}1
    {✗}{{$\times$}}1
    {<}{{\textless}}1
    {>}{{\textgreater}}1
    {|}{{\textbar}}1
    {\&}{{\&}}1,
}
\newtcblisting{promptbox}[1][]{
  enhanced,
  breakable,
  colback=gray!4,
  colframe=gray!55,
  boxrule=0.4pt,
  arc=2pt,
  top=4pt,
  bottom=4pt,
  left=6pt,
  right=6pt,
  listing only,
  listing options={
    language=promptlang,
    basicstyle=\footnotesize\rmfamily,
    numbers=none,
    breaklines=true,
    showstringspaces=false,
    columns=fullflexible,
    keepspaces=true,
    inputencoding=utf8,
    extendedchars=true,
    mathescape=false,
    escapeinside={},
  },
  before skip=6pt,
  after skip=6pt,
  boxsep=2pt,
  #1
}

\definecolor{medgray55}{gray}{0.55}
\definecolor{medgray}{gray}{0.7}
\definecolor{litegray}{gray}{0.9}
\definecolor{gblue}{RGB}{210, 227, 252}
\definecolor{gred}{RGB}{250, 210, 207}
\definecolor{gyellow}{RGB}{254, 239, 195}
\definecolor{ggreen}{RGB}{206, 234, 214}
\definecolor{gorange}{RGB}{254, 223, 200}
\definecolor{gblue9}{RGB}{23, 78, 166}
\definecolor{gred9}{RGB}{165, 14, 14}
\definecolor{gyellow9}{RGB}{227, 116, 0}
\definecolor{ggreen9}{RGB}{13, 101, 45}
\definecolor{gorange9}{RGB}{176, 96, 0}
\definecolor{myblue}{rgb}{0,0,1}
\definecolor{myred}{rgb}{1,0,0}
\definecolor{mylightgray}{gray}{0.95}
\definecolor{myCite}{HTML}{1C4587}
\definecolor{highlightblue}{HTML}{185ABC}
\definecolor{cellHighlight}{HTML}{dbefff}

\newcolumntype{L}[1]{>{\raggedright\let\newline\\\arraybackslash\hspace{0pt}}m{#1}}
\newcolumntype{C}[1]{>{\centering\arraybackslash}m{#1}}
\newcolumntype{R}[1]{>{\raggedleft\let\newline\\\arraybackslash\hspace{0pt}}m{#1}}

\newcommand{\cmark}{\ding{51}}

\let\cite\citep
\hypersetup{
  colorlinks=true,
  citecolor=myCite,
  linkcolor=myCite,
  urlcolor=myCite
}

\providecommand{\equalcontrib}{\textsuperscript{*}}
\providecommand{\corresponding}{\textsuperscript{\dag}}

\title{IACM-RL: Intent-Aware Context Management and Reinforcement Learning for Complex Tool Invocation under Dynamic Intent Fluctuations}
\author{
    Dingwei Zhu\textsuperscript{1}\equalcontrib,
    Jiahan Li\textsuperscript{1}\equalcontrib,
    Chengjun Pan\textsuperscript{3},
    Yunxian Yang\textsuperscript{1},
    Yunbin Zhao\textsuperscript{2}\mbox{,}\\\vspace{-0.2em}\bfseries
    Yunke Zhang\textsuperscript{2},
    Zhonghang Lu\textsuperscript{1},
    Zhuohui Sheng\textsuperscript{1},
    Chenhao Huang\textsuperscript{1},
    Jiahang Lin\textsuperscript{1}\mbox{,}\\\bfseries
    Yajie Yang\textsuperscript{1},
    Junlin Shang\textsuperscript{1},
    Shichun Liu\textsuperscript{1},
    Yuhui Wang\textsuperscript{1},
    Honglin Guo\textsuperscript{1}\mbox{,}\\\bfseries
    Junjie Ye\textsuperscript{1},
    Xin Guo\textsuperscript{1},
    Jiazheng Zhang\textsuperscript{1},
    Ming Zhang\textsuperscript{1},
    Shihan Dou\textsuperscript{1}\mbox{,}\\\bfseries
    Zhiheng Xi\textsuperscript{1},
    Tao Gui\textsuperscript{1}\corresponding,
    Qi Zhang\textsuperscript{1},
    Xipeng Qiu\textsuperscript{1},
    Xuanjing Huang\textsuperscript{1}
    \\\normalfont
    \textsuperscript{1}Fudan University\quad
    \textsuperscript{2}Honor Device Co., Ltd.\quad
    \textsuperscript{3}Peking University\\
    \texttt{dwzhu25@m.fudan.edu.cn, tgui@fudan.edu.cn}
}

\begin{abstract}
Executing long-horizon tool invocations in real-world environments is severely challenged by dynamic user intent noise. Existing methods attempt robustness via implicit history scanning or text compression, yet predominantly assume perfect instructions in simplistic scenarios. Inevitably, under fluctuating contexts, obsolete constraints dilute model attention, triggering intent deviation and repetitive API loops. To resolve this, we propose \textbf{IACM-RL}, a comprehensive framework for robust tool invocation. First, we introduce the \textbf{DynamicIntent} pipeline, synthesizing trajectories across several fine-grained fluctuation scenarios, paired with a five-dimensional diagnostic metric suite. Second, IACM-RL deploys a BeliefState-based Self-Generated Context Manager that proactively tracks shifting goals and isolates overwritten parameters using structural stale flags. To autonomously internalize this state-tracking capability, we optimize the policy using a hierarchical intent-driven reward alongside two auxiliary losses (CM extraction and state distillation). Experiments on DynamicIntent, BFCL-V3, and $\mathrm{\tau}^2$-Bench demonstrate that IACM-RL outperforms  baselines, reducing infinite loops and stale context errors while enhancing out-of-domain generalization.
\end{abstract}

\titlespacing*{\section}{0pc}{2ex plus3pt minus2pt}{4pt}
\titlespacing*{\subsection}{0pc}{1.8ex plus2pt minus1.5pt}{2pt}
\titlespacing*{\subsubsection}{0pc}{1.5ex plus2pt minus1pt}{1pt}
\titlespacing*{\paragraph}{0pc}{1.4ex plus1.5pt minus1pt}{0pt}

\begin{document}

\maketitle
\begingroup
  \renewcommand{\thefootnote}{}
  \footnotetext{\equalcontrib Equal contribution. \corresponding Corresponding author.}
\endgroup

\newcommand{\SkipImportedBibliography}[1]{}

\begingroup
  \let\maketitle\relax
  \RenewEnviron{abstract}{}
  \let\bibliography\SkipImportedBibliography

\maketitle

\begin{abstract}
Executing long-horizon tool invocations in real-world environments is severely challenged by dynamic user intent noise. Existing methods attempt robustness via implicit history scanning or text compression, yet predominantly assume perfect instructions in simplistic scenarios. Inevitably, under fluctuating contexts, obsolete constraints dilute model attention, triggering intent deviation and repetitive API loops. To resolve this, we propose \textbf{IACM-RL}, a comprehensive framework for robust tool invocation. First, we introduce the \textbf{DynamicIntent} pipeline, synthesizing trajectories across several fine-grained fluctuation scenarios, paired with a five-dimensional diagnostic metric suite. Second, IACM-RL deploys a BeliefState-based Self-Generated Context Manager that proactively tracks shifting goals and isolates overwritten parameters using structural stale flags. To autonomously internalize this state-tracking capability, we optimize the policy using a hierarchical intent-driven reward alongside two auxiliary losses (CM extraction and state distillation). Experiments on DynamicIntent, BFCL-V3, and $\mathrm{\tau}^2$-Bench demonstrate that IACM-RL outperforms  baselines, reducing infinite loops and stale context errors while enhancing out-of-domain generalization.
\end{abstract}

\begin{figure}[htb]
\centering
\includegraphics[width=0.82\linewidth]{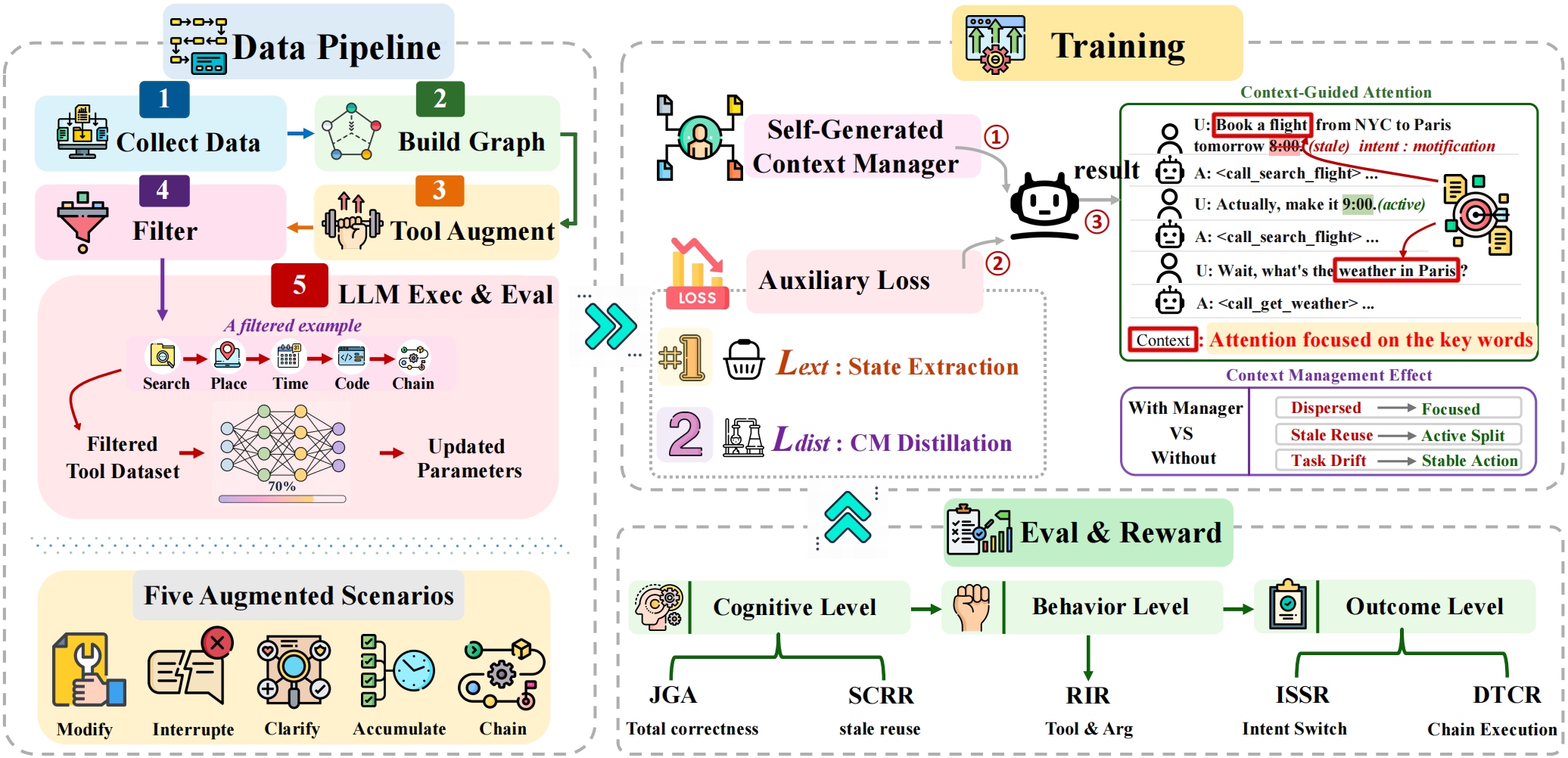}
\caption{Overview of the IACM-RL framework. (Left) The scalable DynamicIntent data pipeline that synthesizes multi-turn trajectories across five intent-fluctuation modes. (Top Right) The training paradigm, featuring an autoregressive BeliefState-based Context Manager that proactively tracks shifting goals and utilizes structural stale flags to decouple state tracking from action generation. (Bottom Right) The optimization process, which internalizes state-tracking capabilities using a hierarchical intent-driven reward system (spanning cognitive, behavioral, and outcome dimensions) alongside two auxiliary losses.}
\label{fig:main}
\end{figure}

\section{Introduction}

Large Language Models (LLMs)~\cite{zhu2026dvpodistributionalvaluemodelingbased,zhu-etal-2026-vrpo,zhu2026dfposcalingvaluemodeling,xi-etal-2026-agentgym2,pan2026evpoexplainedvariancepolicy} have demonstrated profound potential in automated tool invocation, yet existing training paradigms predominantly assume ``perfect static instructions,'' presuming that user queries evolve linearly without interruption.~\cite{zhou2024webarenarealisticwebenvironment,liu2025agentbenchevaluatingllmsagents} In stark contrast, real-world interactions are fraught with intent noise, where users frequently modify tasks, insert irrelevant chit-chat, or articulate ambiguous constraints.~\cite{3618408.3619699,li-etal-2024-evaluating-instruction} Achieving robust reinforcement learning (RL) under such noisy supervision remains a critical, unresolved prerequisite for deploying reliable AI agents.

To mitigate these dynamic perturbations, previous methodologies attempt robustness through long-context fine-tuning, implicit reasoning chains, or context compression~\cite{kang2026aconoptimizingcontextcompression}. However, these paradigms predominantly rely on the assumption of perfect instructions, forcing agents to implicitly scan verbose, raw histories. Consequently, under complex and fluctuating contexts, the accumulation of obsolete constraints and irrelevant interruptions severely dilutes the model's token-level attention.~\cite{liu-etal-2024-lost} Critical tokens representing updated goals receive disproportionately low attention weights, causing the agent to overlook key information. This severe attention dilution subverts multi-tool data dependencies, inevitably trapping the agent in catastrophic intent deviation and infinite API calling loops~\cite{11391755,arike2025technicalreportevaluatinggoal}. While Contextual Belief Management tracks evidence shifts under contextual pressure~\cite{xu2026modelschangemindscontextual}, and probing studies underscore the necessity of explicit external guidance due to LLMs' unstable internal representations~\cite{luo2026probinglackstableinternal}, these approaches exhibit fault-tolerance only in simplistic tool scenarios and collapse when confronted with long-horizon, parallel multi-tool tasks.

To resolve this, we propose \textbf{IACM-RL} (Figure~\ref{fig:main}), a comprehensive framework for robust tool invocation. Concretely, IACM-RL integrates the \textbf{DynamicIntent} pipeline, a scalable data construction framework that synthesizes multi-turn trajectories spanning several fine-grained intent fluctuation scenarios, accompanied by a five-dimensional metric suite for dynamic intent diagnosis. At its core, a BeliefState-based Self-Generated Context Manager neutralizes history-induced noise by proactively tracking shifting goals and isolating overwritten parameters via structural stale flags, forcibly decoupling critical states from verbose histories. Crucially, IACM-RL autonomously internalizes this mechanism. Guided by a hierarchical intent-driven reward and two auxiliary losses (CM extraction and state distillation), the agent learns to dynamically extract robust state blocks from noisy trajectories and distill this explicit tracking capability directly into its implicit parameters.

We validate IACM-RL on the DynamicIntent Benchmark encompassing both ID and OOD scenarios, alongside BFCL-V3 and $\mathrm{\tau}^2$-Bench. Empirical results demonstrate that IACM-RL achieves the highest overall average among all methods, demonstrates strong out-of-domain generalization, and shows consistent gains on adversarial drift and per-scenario analysis.
Our core contributions are summarized as follows:

\noindent\textbf{DynamicIntent Dataset and Benchmark:} We construct a large-scale dataset of multi-turn tool-calling trajectories spanning several fine-grained intent fluctuation scenarios, and establish the DynamicIntent Benchmark with a five-dimensional metric suite for dynamic intent diagnosis, covering both ID and OOD evaluation.

\noindent\textbf{Self-Generated Context Manager via BeliefState:} We introduce an explicit Context Manager where the model proactively tracks user goals and marks overwritten parameters with a structural stale flag, targeting dynamic user intent scenarios.

\noindent\textbf{Intent-Driven RL Optimization:} We design a hierarchical reward system providing dense supervision across cognitive, behavioral, and outcome dimensions. Crucially, two auxiliary losses empower the agent to autonomously internalize robust, long-horizon state tracking.

\begin{figure}[htb]
\centering
\includegraphics[width=0.80\linewidth]{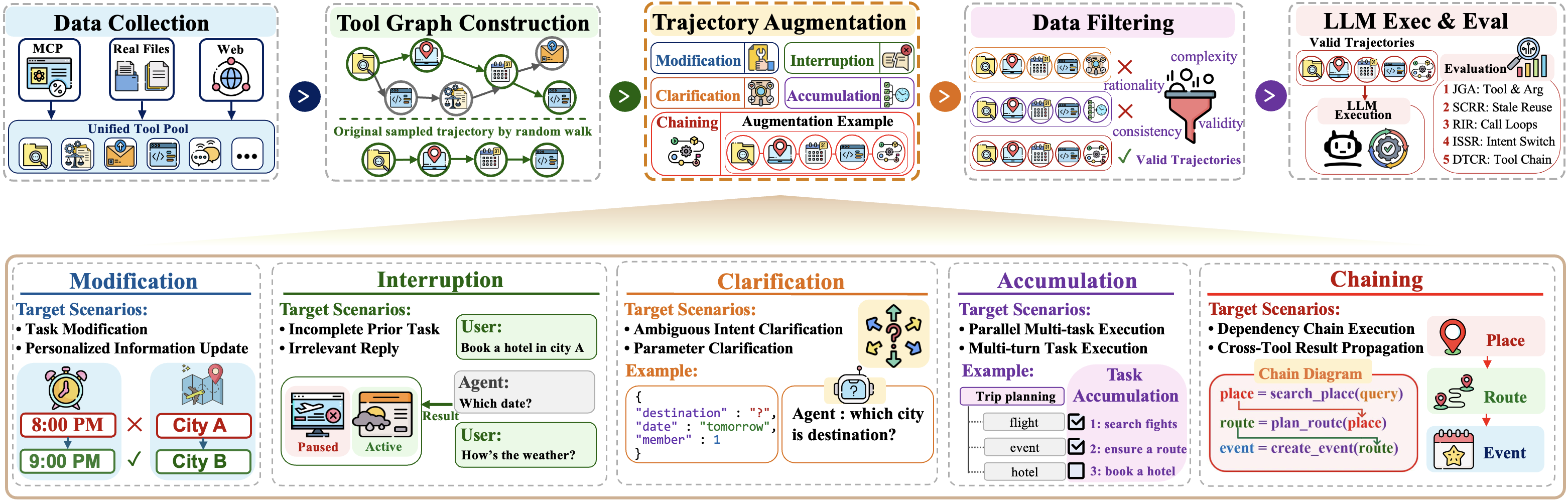}
\caption{DynamicIntent data construction pipeline. Complex trajectories are synthesized through (1) unified tool abstraction across heterogeneous APIs, (2) dependency graph construction via constraint-aware LLM voting, (3) trajectory augmentation injecting realistic behavioral noise across several fine-grained scenarios (grouped into Modification, Interruption, Clarification, Accumulation, and Chaining modes), (4) multi-dimensional filtering for consistency, plausibility, and complexity, and (5) LLM-driven simulation execution.}
\label{fig:pipeline}
\end{figure}

\section{Related Work}

\noindent\textbf{Intent Fluctuation and Contextual Belief Management.} Intent fluctuation describes how an agent's objectives deviate amid dynamic user feedback and redundant context, fundamentally challenging the LLM assumption of static instruction adherence. Precise intent tracking strips away historical noise and captures genuine requirement shifts, thereby improving multi-turn tool stability. Early works quantify this deviation by tracking temporal transitions~\cite{11391755}, observing goal divergence as contextual signals accumulate~\cite{arike2025technicalreportevaluatinggoal}, and designing intent assurance frameworks to extract key indicators for drift detection~\cite{dzeparoska2024intentassuranceusingllms}. Recently, \cite{xu2026modelschangemindscontextual} extended dynamic tracking to explicit belief state maintenance to capture evidence shifts, while other studies examined how explicit value conflicts trigger asymmetric goal drift~\cite{saebo2026asymmetricgoaldriftcoding}. Furthermore, \cite{luo2026probinglackstableinternal,sun2026evolutionarycontextsearchautomated,kang2026aconoptimizingcontextcompression} revealed that LLMs lack stable internal representations, necessitating external belief guidance. Building upon these insights, our IACM-RL framework autonomously manages dynamic intents by internalizing a self-generated BeliefState directly within the RL optimization loop.

\noindent\textbf{Multi-Turn Reasoning Instability and Agent State Management.} Maintaining reasoning stability in long-horizon environments is a critical challenge, as LLMs frequently lose relevant evidence, succumb to noise, exhibit contextual inertia, and suffer instruction degradation~\cite{liu-etal-2024-lost,3618408.3619699,chen2026breakingcontextualinertiareinforcement,laban2025llmslostmultiturnconversation}, often leading to severe error cascades across execution steps~\cite{ma2026signaldifferentsemanticscrossframework,lin2026agentaskmultiagentsystemsneed}. To mitigate this, recent frameworks utilize reinforcement learning to jointly optimize memory~\cite{zhou2025mem1learningsynergizememory}, refine prompts to filter noise~\cite{kang2026aconoptimizingcontextcompression}, or employ scalable external memory layers to consolidate salient information across sessions~\cite{chhikara2025mem0buildingproductionreadyai}. Unlike these methods, IACM-RL empowers the agent to autonomously internalize complex state management by explicitly decoupling state tracking from action generation via a self-generated BeliefState.

\section{Intent-Aware Tool Dataset and \\ Evaluation Framework}

The scarcity of high-quality, multi-turn dialogue trajectories containing complex intent fluctuations constitutes a foundational bottleneck in robust tool-calling research. Because existing benchmarks predominantly focus on static, single-shot instructions, we introduce the \textbf{DynamicIntent Dataset}. This dataset is generated via a highly scalable simulation pipeline and is accompanied by a novel multi-dimensional evaluation metric suite specifically designed for dynamic intent diagnosis. The remainder of this section details the construction pipeline, the fluctuation modes, and the resulting corpus statistics.

\subsection{Data Construction Pipeline}
Our pipeline synthesizes complex multi-turn tool-calling trajectories without real backends, in five stages (Figure~\ref{fig:pipeline}).

\noindent\textbf{Tool Abstraction.} We aggregate heterogeneous APIs (English tool sets plus proprietary third-party and mid-control applications) and normalize each into a unified schema covering description, parameters, and response, so that generalization pressure falls on schema reading rather than name memorization.

\noindent\textbf{Dependency Graph.} We build a directed Tool Dependency Graph in which an edge indicates that the return of one tool can semantically satisfy a parameter of another. Edges are established by multi-source LLM voting with cross-checks for precision, and constraint-aware random walks on the graph sample baseline Function Sequence Patterns (FSP) as workflow backbones.

\noindent\textbf{Trajectory Augmentation.} Intent-augmentation methods inject realistic behavioral noise into the FSP backbones across five modes that map to the BeliefState fields (Section~\ref{sec:aug_by_mode}):\textbf{ Modify} overwrites set parameters; \textbf{Interrupt} splices an orthogonal sub-trajectory under a tool-conflict check so the new task is genuinely independent; \textbf{Clarify} omits mandatory parameters or prepends fragmented intents to force proactive questioning; \textbf{Accumulate} appends parallel or inherited requests; and\textbf{ Chaining} extends the FSP along dependency edges into long tool-return-to-parameter chains.

\noindent\textbf{Data Filtering.} Each trajectory is screened for consistency (schema-conformant calls whose arguments match the declared parameters and correctly map user-mentioned information), plausibility (a closed loop that resolves the request, with coherent returns and summaries), and complexity (a length budget that removes overlong or degenerate samples). The human-annotated ID/OOD test set additionally undergoes a two-tier annotation pass to guarantee exact evaluation labels.

\noindent\textbf{Simulation Execution.} An LLM simulator interprets the filtered FSP traces: it renders natural user queries, synthesizes mock returns conforming to each tool's response schema, and produces assistant summaries, yielding complete dialogues without any real backend. The expected final tool dependencies and argument states are recorded as a golden map that serves as the exact optimization and evaluation target.

\subsection{Detailed Pipeline Stages}
\label{sec:pipeline_detail}

This subsection expands the five-stage pipeline sketched in Section~3.1 with the procedural details omitted from the main text.

\noindent\textbf{Stage 1: Tool Abstraction and Collection.} We aggregate heterogeneous API sources, including English tool sets and proprietary third-party and mid-control applications, and normalize each API into the unified schema via UTAL, so that format differences across sources are hidden from downstream stages. The normalized tools form a cross-source pool from which each trajectory samples its own candidate-tool subset; because the subset varies across samples, the model is forced to read the schema and select the correct tool rather than memorize tool names.

\noindent\textbf{Stage 2: Dependency Graph Construction.} We build the directed Tool Dependency Graph $G$ by asking multiple LLMs, each voting twice (once from schema descriptions alone and once with a concrete input example), whether the return of one tool can satisfy a parameter of another. An edge is kept only when both votes of an account agree and a majority of accounts concur, which suppresses spurious edges and keeps the graph precise. Constraint-aware random walks on $G$ then sample baseline Function Sequence Patterns (FSP) as workflow backbones, ensuring that adjacent tools in a backbone are genuinely related by data flow.

\noindent\textbf{Stage 3: Trajectory Augmentation.} The intent-augmentation methods inject realistic human behavioral noise into the FSP backbones. They split into graph-level methods, which edit the FSP structure over $G$, and trajectory-level methods, which rewrite already-synthesized dialogues, organized into the five intent-fluctuation modes detailed in Section~\ref{sec:aug_by_mode}. A shared subroutine across the interruption-family methods is graph-constrained splicing (Algorithm~\ref{alg:splice}): an orthogonal sub-trajectory is spliced into a base trajectory only when its tools are name-disjoint from the base and no dependency edge connects the two tool sets, guaranteeing that the spliced segment introduces a genuinely new task rather than a logical continuation.

\noindent\textbf{Stage 4: Data Filtering.} Each auto-generated trajectory is screened along three dimensions before RL use:
\begin{itemize}[leftmargin=*,noitemsep,topsep=2pt]
    \item Consistency. Every tool call must be schema-conformant: arguments use the declared parameter names, include all required parameters, contain no undeclared extra keys, respect declared types and allowed values, and correctly map any user-mentioned information that corresponds to a tool parameter. Calls that reference tools absent from the candidate pool are discarded, and recoverable cases are repaired in place.
    \item Plausibility. The trajectory must form a closed loop that resolves the user's request: tool returns must be coherent with the call arguments, and the assistant summary must faithfully reflect the returns. Samples whose task goal diverges substantially from the executed result are deleted, while those needing only a minor parameter fix are corrected.
    \item Complexity. Trajectories are filtered against a prompt-length budget so that the distribution stays informative: overlong samples are clipped or rejected, and trivially short or degenerate samples are removed.
\end{itemize}
The human-annotated ID/OOD test set additionally undergoes a two-tier annotation pass, with universal rules and scenario-specific checks, to guarantee exact evaluation labels.

\noindent\textbf{Stage 5: LLM-Driven Simulation Execution.} An LLM-based simulator interprets the filtered FSP traces via back-and-forth translation: for each tool call it renders a natural user query from the schema, generates the assistant tool call, synthesizes a mock API return conforming to the tool's response schema, and produces an assistant summary, yielding complete multi-turn dialogues without any real backend. Where real responses are available they are cached and reused; otherwise the simulator generates schema-conformant mock returns. The pipeline also records the expected final tool dependencies and argument states as a golden map, which serves as the exact optimization and evaluation target for the rule-based reward and metrics.

\subsection{Augmentation Operators by Intent Mode}
\label{sec:aug_by_mode}

The intent-augmentation methods introduced in the main pipeline (Section~3.1) are organized around the same five intent-fluctuation modes that the BeliefState fields and reward metrics target. Each mode is realized by a combination of graph-level methods and trajectory-level methods. Table~\ref{tab:aug_by_mode} summarizes the mapping; we detail each mode below.

\begin{table}[htb]
\centering
\footnotesize
\begin{tabular}{p{0.11\textwidth}p{0.29\textwidth}p{0.34\textwidth}p{0.12\textwidth}}
\toprule
\textbf{Intent Mode} & \textbf{Graph-level methods} & \textbf{Trajectory-level methods} & \textbf{BeliefState field} \\
\midrule
Modification & --- & ``task change'', ``rewrite param'', ``rewrite function'', ``personalization'' & $\mathcal{C}_{\text{slots}}$ (\texttt{stale}) \\
Interruption & --- & ``function switch'', ``function insert'', ``add fsp'', ``add question node'', ``pre-task new query'' & $\mathcal{G}_{\text{current}}$, $\mathcal{I}_{\text{signal}}$ \\
Clarification & ``FSP split param'', ``FSP split function'' & ``none function'', ``non query'', ``ask intent'', ``clarify unclear intent'' & $\mathcal{Q}_{\text{pending}}$ \\
Accumulation & ``FSP merge'' & ``function parallel'', ``inherit param'', ``inherit function'', ``param reference'' & $\mathcal{C}_{\text{slots}}$, $\mathcal{G}_{\text{current}}$ \\
Chaining & ``FSP insert'', ``FSP insert long'' & --- & $\mathcal{A}_{\text{last}}$ \\
\bottomrule
\end{tabular}
\caption{Augmentation methods grouped by the five intent-fluctuation modes. Graph-level methods act on the FSP over $G$; trajectory-level methods act on synthesized dialogues. Each mode maps to the BeliefState field it most stresses (Eq.~\eqref{eq:beliefstate}).}
\label{tab:aug_by_mode}
\end{table}

\noindent\textbf{Modification.} The user overwrites a previously set parameter, and the agent must not regress to the stale value. The task change method uses an LLM to inject a three-message block at a chosen insertion point: a user message overwriting a parameter, a re-invoked tool call with updated arguments, and a modified tool return. The rewrite param and rewrite function methods paraphrase a parameter value or swap to a functionally similar tool. The personalization method injects a user profile that is updated between rounds so the same tool yields different arguments. These directly exercise the stale flag on $\mathcal{C}_{\text{slots}}$.

\noindent\textbf{Interruption.} The user switches to a new task mid-flow. All five methods splice an orthogonal sub-trajectory into a base trajectory under the graph-constrained conflict check of Algorithm~\ref{alg:splice}, differing only in insertion position. Function switch appends the new task to the end of the trajectory. Function insert splices 1--3 rounds into the middle. Pre-task new query inserts after the last user or last tool message, yielding the after-user and after-tool variants. Add fsp and add question node attach a new complex or simple task right after a parameter-clarification question. They test $\mathcal{G}_{\text{current}}$ and the interrupted $\mathcal{I}_{\text{signal}}$.

\noindent\textbf{Clarification.} Information is missing or ambiguous, forcing proactive questioning. The graph-level FSP split param method marks a parameter as intentionally omitted, and FSP split function marks a missing tool; during trajectory synthesis these flags trigger branches where the assistant must ask for the missing information. The trajectory-level none function and non query methods insert empty or chitchat turns, while ask intent and clarify unclear intent prepend LLM-generated ambiguous or fragmented utterances. These populate $\mathcal{Q}_{\text{pending}}$.

\noindent\textbf{Accumulation.} The user appends parallel or inherited requests. The graph-level FSP merge method merges two FSPs so the agent executes two tool chains in one turn. The trajectory-level function parallel method appends a second request to an existing user turn. Inherit param, inherit function, and param reference replace a turn with a structurally similar tool whose arguments are inherited from prior results rather than re-stated by the user. These stress multi-goal slot management in $\mathcal{C}_{\text{slots}}$ and $\mathcal{G}_{\text{current}}$.

\noindent\textbf{Chaining.} The return of tool A feeds the parameter of tool B, so prior results must be retained. This mode is realized purely at the graph level. The FSP insert method appends a graph-successor along a dependency edge, and FSP insert long enforces a non-empty dependency to build a deeper chain. The resulting trajectories rely on $\mathcal{A}_{\text{last}}$ to chain calls correctly.

\subsection{Per-Scenario Construction Mechanisms}

Beyond the five-stage pipeline described in the main text, each fine-grained intent fluctuation scenario is instantiated by a dedicated trajectory-transformation method that injects a specific type of intent noise into a base FSP trace. These methods fall into three families: graph-constrained splicing (composing two independent trajectories under a tool-conflict check), LLM-driven rewriting (regenerating part of the dialogue to inject a perturbation), and prefix augmentation (prepending a clarification sub-dialogue). We summarize the construction mechanism of each scenario below.

\noindent\textbf{Baseline scenarios.} Simple Task (Single-turn) and Simple Task (Multi-turn) are produced directly by the base FSP random walk and back-and-forth translation, with no augmentation method applied; they serve as the non-fluctuation reference.

\noindent\textbf{Modification scenarios.} Task Modification uses an LLM to rewrite a complete trajectory: given a finished dialogue, the LLM identifies an insertion point and injects a three-message block, namely a user message overwriting a previously set parameter, a tool call re-invoking the same tool with updated arguments, and a tool message returning the modified result. This is the scenario that most directly exercises the structural stale flag, as the overwritten value must not be reused. Because modification samples are naturally rare, this category is deliberately over-augmented to 258 trajectories. Personalized Information Update injects a user profile and device snapshot before a user request, and the profile is updated across two rounds so that re-invoking the same tool yields different argument values, verifying that the agent tracks evolving personal context rather than caching stale parameters.

\noindent\textbf{Interruption scenarios.} Incomplete Prior Task (After User) and Incomplete Prior Task (After Tool) both insert a new query at different positions: in the After-User variant the new query interrupts immediately after the user's instruction, before the assistant has acted; in the After-Tool variant the interruption occurs after a tool has returned but before the assistant summarizes, so the agent must abandon the pending summary and pivot to the new goal. In both cases the spliced sub-trajectory is selected under a tool-conflict and graph-edge check: the candidate's tools must not overlap with the already-used tools, and no dependency edge may exist between them, ensuring the new task is genuinely orthogonal rather than a logical continuation. Irrelevant Reply (Simple Task) and Irrelevant Reply (Complex Task) occur when the assistant issues a parameter-clarification question and the user replies with an unrelated simple or complex request, testing whether the agent can abandon the suspended clarification and service the new request.

\noindent\textbf{Clarification scenarios.} Parameter Clarification purposefully omits a mandatory parameter from the user query, so the assistant must proactively ask for it before invoking the tool. Ambiguous Intent Clarification prepends a deliberately fragmented user utterance that is neither chitchat nor a complete request, forcing the assistant to clarify the true intent before the original task proceeds. Multi-turn Recursive Clarification stacks an additional intent-clarification round on top of parameter clarification. Linguistic Clarification denotes the pure-language subset that is mixed in to prevent conversational-capability forgetting.

\noindent\textbf{Accumulation scenario.} Parallel Multi-task Execution merges two independent FSP traces into a single trajectory where the agent must execute two orthogonal tool chains within one response, testing concurrent multi-goal management.

\noindent\textbf{Chaining scenario.} Dependency Chain Execution extends a base FSP by inserting graph-successor tools along dependency edges, forcing the agent to consume the return value of one tool as the input of the next, providing the structural basis for the DTCR metric.

\noindent\textbf{Graph-constrained splicing.} The interruption and irrelevant-reply scenarios share a common subroutine: splicing an independent sub-trajectory $B$ into a base trajectory $A$. To guarantee that $B$ introduces a genuinely orthogonal task rather than a logical continuation, we apply a two-level conflict check against the tool dependency graph $G$ before accepting a candidate $B$: the set of tool names invoked in $B$ must not intersect those already used in $A$, and no directed dependency edge may exist in $G$ between any tool used in $A$ and any tool used in $B$. Only a candidate satisfying both conditions is spliced, after which round identifiers and tool-call identifiers are re-indexed and the candidate tool pool is de-duplicated. Algorithm~\ref{alg:splice} summarizes this procedure.

\subsection{Overall Composition and Scale}
\label{sec:scenario_dist}

The finalized corpus contains four splits. The multi-turn training set consists of 5,639 trajectories; the single-step training set contains 17,391 trajectories. To prevent catastrophic forgetting of conversational capabilities during RL, approximately $9.3\%$ of the multi-turn set and $9.5\%$ of the single-step set are supplemented with pure linguistic clarification queries. Table~\ref{tab:dataset_overview} summarizes the scale, average tool-pool size, and expected target calls per trajectory.

\begin{table}[htb]
\centering
\small
\begin{tabular}{lcccc}
\toprule
\textbf{Dataset Split} & \textbf{Total Samples} & \textbf{Mean Tools in Pool} & \textbf{Max Tools}  \\
\midrule
Multi-turn Train       & 5,639  & 11.7 & 40  \\
Multi-turn Validation  & 753    & ---  & --- \\
Single-step Train      & 17,391 & 12.2 & 40  \\
Single-step Validation & 3,733  & ---  & --- \\
\bottomrule
\end{tabular}
\caption{Overview of the DynamicIntent Dataset splits.}
\label{tab:dataset_overview}
\end{table}

\begin{table}[htb]
\centering
\footnotesize
\setlength{\tabcolsep}{4pt}
\begin{tabular}{@{}lrrrrrrc@{}}
\toprule
\multirow{2}{*}{\textbf{Fine-Grained Scenario}} & \multicolumn{2}{c}{\textbf{Multi-turn Train}} & \multicolumn{2}{c}{\textbf{Single-step Train}} & \multicolumn{2}{c}{\textbf{Test Dialogs}} & \multirow{2}{*}{\textbf{Mode}} \\
\cmidrule(lr){2-3} \cmidrule(lr){4-5} \cmidrule(lr){6-7}
 & Count & \% & Count & \% & ID & OOD & \\
\midrule
Parameter Clarification            & 1,052 & 18.7 & 1,910 & 11.0 & 100 & 100 & Clarification \\
Incomplete Prior Task (After User) & 986   & 17.5 & 3,681 & 21.2 & 100 & 100 & Interruption \\
Parallel Multi-task Execution      & 956   & 17.0 & 2,250 & 12.9 & 100 & 100 & Accumulation \\
Dependency Chain Execution         & 617   & 10.9 & 725   & 4.2  & 100 & 100 & Chaining \\
Linguistic Clarification           & 523   & 9.3  & 1,650 & 9.5  & --- & --- & Clarification \\
Simple Task (Single-turn)          & 408   & 7.2  & 3,289 & 18.9 & 100 & 78  & Baseline \\
Simple Task (Multi-turn)           & 390   & 6.9  & 2,093 & 12.0 & 100 & 100 & Baseline \\
Ambiguous Intent Clarification     & 350   & 6.2  & 817   & 4.7  & 100 & 100 & Clarification \\
Task Modification                  & 258   & 4.6  & 590   & 3.4  & 100 & 100 & Modification \\
Multi-turn Recursive Clarification & 60    & 1.1  & 146   & 0.8  & 100 & 100 & Clarification \\
Irrelevant Reply (Simple Task)     & 32    & 0.6  & 96    & 0.6  & 100 & 100 & Interruption \\
Irrelevant Reply (Complex Task)    & 4     & 0.1  & 43    & 0.2  & 100 & 100 & Interruption \\
Incomplete Prior Task (After Tool) & 3     & 0.1  & 38    & 0.2  & 100 & 100 & Interruption \\
Personalized Information Update    & ---   & ---  & 63    & 0.4  & 72  & 100 & Modification \\
\midrule
\textbf{Total} & \textbf{5,639} & \textbf{100} & \textbf{17,391} & \textbf{100} & \textbf{1,272} & \textbf{1,278} & \\
\bottomrule
\end{tabular}
\caption{Per-scenario composition of the DynamicIntent Dataset. ``Multi-turn Train'' ($N=5{,}639$) and ``Single-step Train'' ($N=17{,}391$) give the training counts; ``Test Dialogs'' gives the human-annotated ID/OOD test sizes (1,272 and 1,278). ``---'' marks scenarios absent from a split; modes follow Table~\ref{tab:aug_by_mode}.}
\label{tab:scenario_distribution}
\end{table}

\subsection{Per-Trajectory Tool Statistics}

Figure~\ref{fig:tool_dist} shows the distribution of optimal tool count and candidate tool pool size for both the single-step and multi-turn sets. All tools in the dataset are text-based API calls; no multimodal tools are included. In the single-step set, most trajectories ($71.1\%$) require only 1 tool (mean 1.1), while the multi-turn set shifts toward 2 tools ($56.1\%$, mean 1.9), reflecting more complex multi-tool workflows. The candidate tool pool is similar across both sets: single-step averages 12.2 tools (p50 at 8, p90 at 32), multi-turn averages 11.7 (p50 at 6, p90 at 32), ensuring sufficient selection pressure for schema-reading rather than name memorization.

\begin{figure}[htb]
\centering
\includegraphics[width=0.82\linewidth]{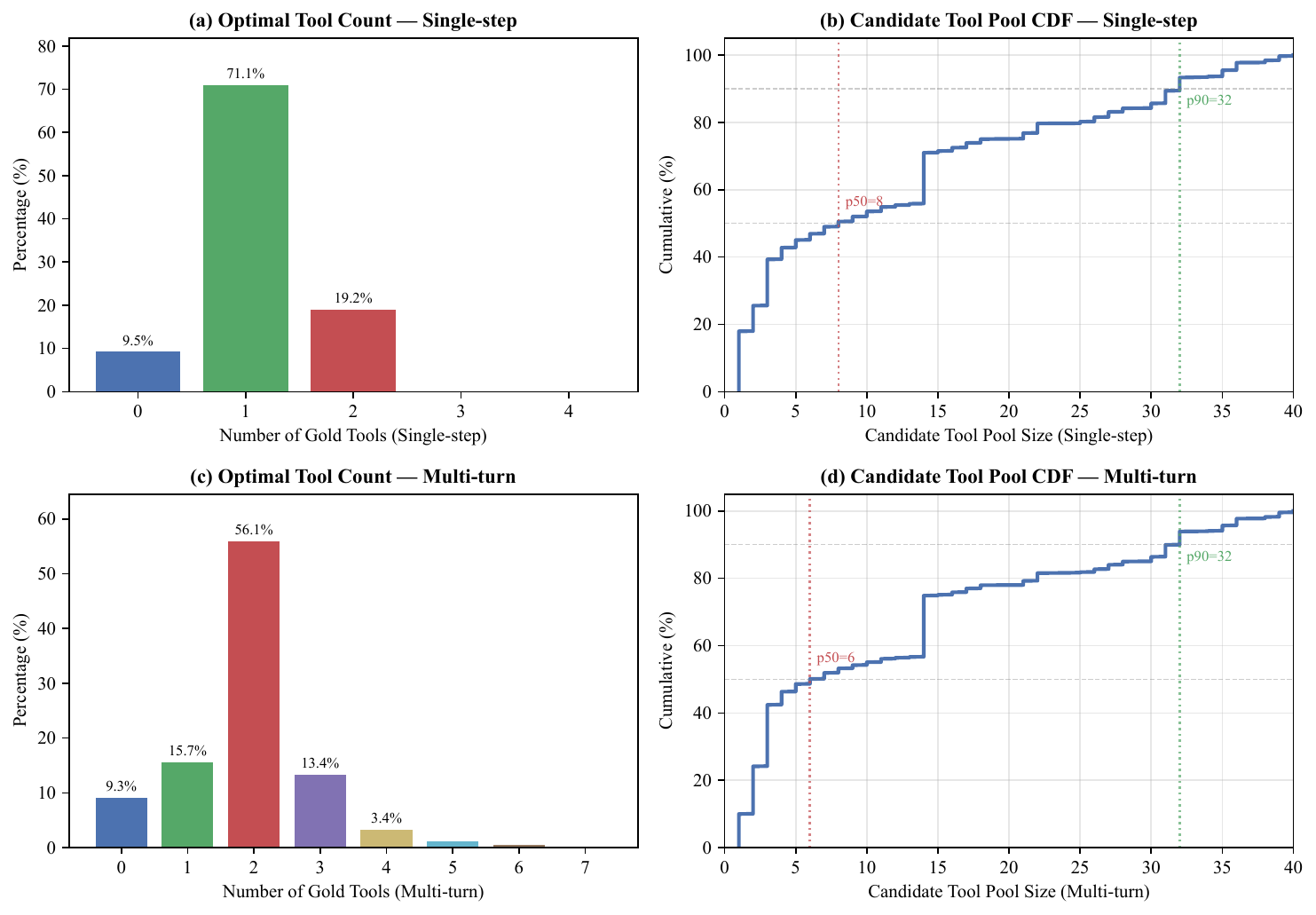}
\caption{Tool count distributions for single-step (top) and multi-turn (bottom) training sets. Left: optimal tools per trajectory. Right: candidate tool pool size CDF.}
\label{fig:tool_dist}
\end{figure}

\subsection{Tool Category Classification}

The 1,896 unique tools across both training sets were classified into 9 functional categories via an automated keyword-based classifier that inspects each tool's name and description. The categories, ordered by prevalence in the single-step set, are: Open/Close (e.g., toggle switches, engine start/stop, lock/unlock, dismiss alarms), OCR/Document (file system operations, notes, printing), Navigation (maps, flights, transportation, reservation management), Finance (stock trading, investment, portfolio), Communication (calls, messaging, notifications), Payment (orders, checkout, currency exchange, refunds), Query (information retrieval, weather, calculation, schedule reading), Set/Configure (settings, alarm/reminder editing, app preference management), and Media (audio/video playback, camera, screenshots).

Table~\ref{tab:tool_categories} reports the category distribution for both training sets, counting each tool appearance in the candidate tool pool of every sample. The total counts (212,901 single-step, 65,927 multi-turn) reflect that each sample carries approximately 12 candidate tools on average. The distribution is broadly consistent across the two sets, with Open/Close, OCR/Document, Navigation, and Finance together covering roughly half of all tool occurrences. The ``Other'' category (30.3\% single-step, 29.5\% multi-turn) primarily consists of tools with vague or compound descriptions, many from the vehicle-control and smart-device domains, that resist unambiguous single-category assignment.

\begin{table}[htb]
    \centering
    \small
    \begin{tabular}{lrrrr}
    \toprule
    & \multicolumn{2}{c}{\textbf{Single-step}} & \multicolumn{2}{c}{\textbf{Multi-turn}} \\
    \cmidrule(lr){2-3} \cmidrule(lr){4-5}
    \textbf{Tool Category} & Count & \% & Count & \% \\
    \midrule
    Open/Close      & 30{,}892 & 14.5 & 9{,}767  & 14.8 \\
    OCR/Document    & 27{,}623 & 13.0 & 7{,}728  & 11.7 \\
    Navigation      & 23{,}844 & 11.2 & 9{,}150  & 13.9 \\
    Finance         & 17{,}979 & 8.4  & 4{,}281  & 6.5 \\
    Communication   & 14{,}899 & 7.0  & 4{,}530  & 6.9 \\
    Payment         & 12{,}326 & 5.8  & 3{,}763  & 5.7 \\
    Query           & 10{,}458 & 4.9  & 3{,}867  & 5.9 \\
    Set/Configure   & 5{,}360  & 2.5  & 1{,}598  & 2.4 \\
    Media           & 5{,}100  & 2.4  & 1{,}773  & 2.7 \\
    Other           & 64{,}420 & 30.3 & 19{,}470 & 29.5 \\
    \midrule
    Total           & 212{,}901 & 100.0 & 65{,}927 & 100.0 \\
    \bottomrule
    \end{tabular}
    \caption{Tool category distribution for both training sets. Each tool in the candidate pool of every sample is counted, so counts represent total tool-pool instances. Tools are classified into 9 functional categories via an automated keyword-based classifier; tools not fitting any category are assigned to ``Other.'' The ``Other'' category is dominated by tools with vague or compound descriptions that cannot be unambiguously assigned. All tools are text-based API calls; the dataset does not include multimodal tools.}
    \label{tab:tool_categories}
\end{table}

\begin{figure}[htb]
\centering
\includegraphics[width=0.82\textwidth]{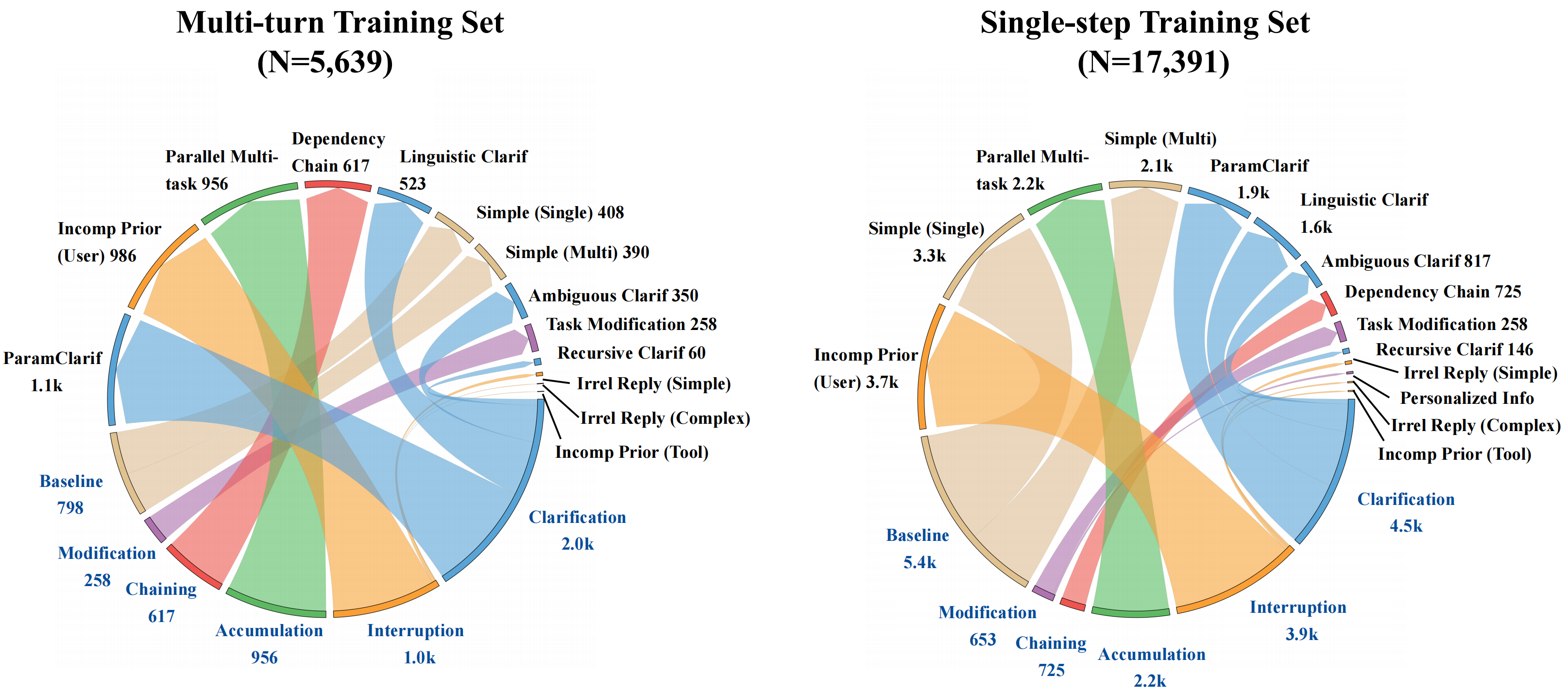}
\caption{Per-scenario distribution of both training sets as chord diagrams. Left: multi-turn ($N{=}5{,}639$); Right: single-step ($N{=}17{,}391$). Each sector is colored by its intent-fluctuation mode, with labels showing count.}
\label{fig:scn_chord}
\end{figure}

\begin{figure}[htb]
\centering
\includegraphics[width=0.82\textwidth]{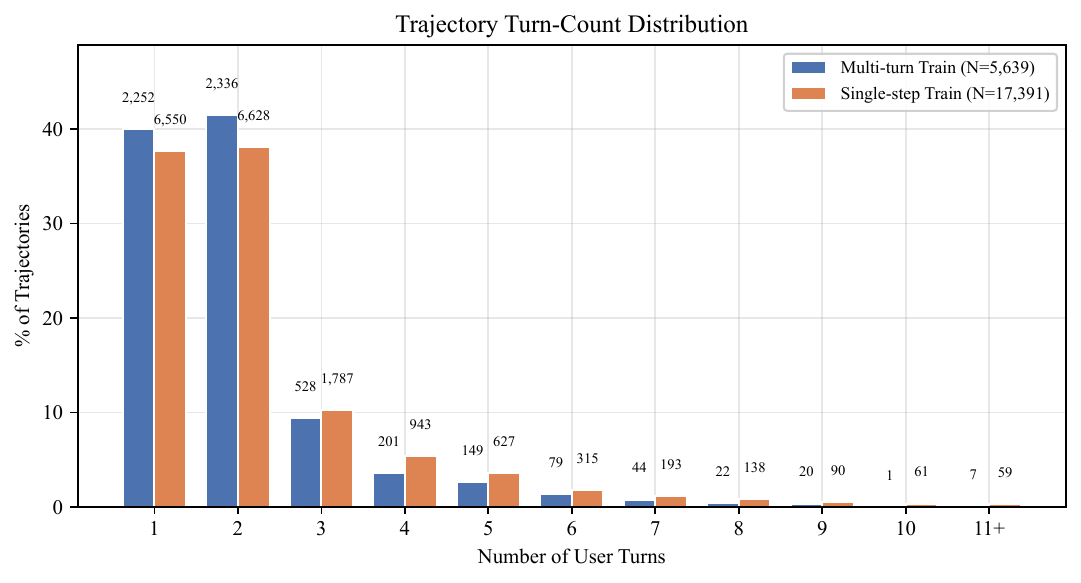}
\caption{Trajectory turn-count distribution. The raw corpus contains trajectories of up to 17 user turns. Before Stage 2 training, trajectories exceeding 12 user/assistant turns are filtered out. The reported Stage 2 training count is computed after this filtering.}
\label{fig:turn}
\end{figure}

\begin{figure}[htb]
\centering
\includegraphics[width=0.82\textwidth]{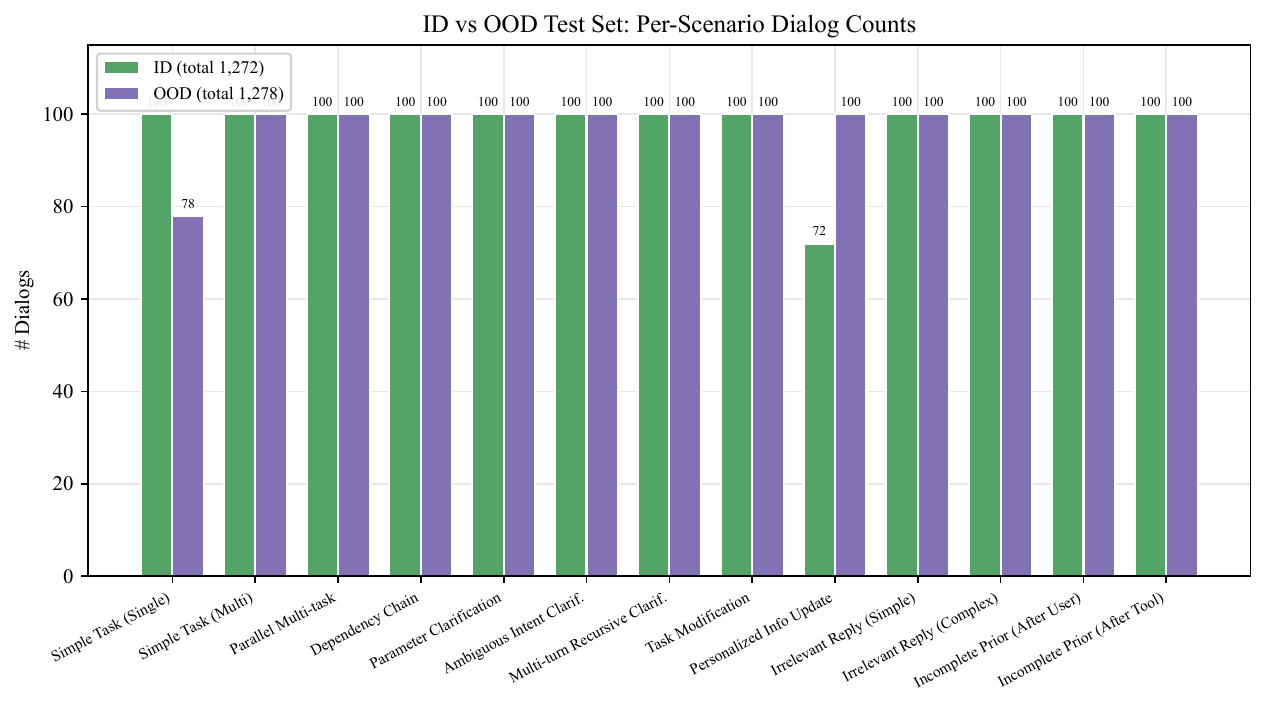}
\caption{Per-scenario dialog counts of the ID and OOD test sets. Scenario coverage is matched; per-scenario volume is comparable.}
\label{fig:id_ood}
\end{figure}

\subsection{Prompt Token-Length Distribution}

Figure~\ref{fig:token_cdf} shows the cumulative distribution of prompt length for the two training sets. The dashed lines mark exact char percentiles after overlong filtering: the single-step set has p50 of 2,387, p90 of 5,076, p95 of 5,887, and p99 of 11,961; the multi-turn set has a tighter p99 of 8,698. Both distributions are right-skewed, with the single-step set exhibiting a heavier tail from the mixed open-source trajectories. The few overlong samples in the single-step set are handled by left-truncation preserving the most recent messages, while the multi-turn set uses error-mode truncation to avoid corrupting structured state.

\begin{figure}[htb]
\centering
\includegraphics[width=0.82\textwidth]{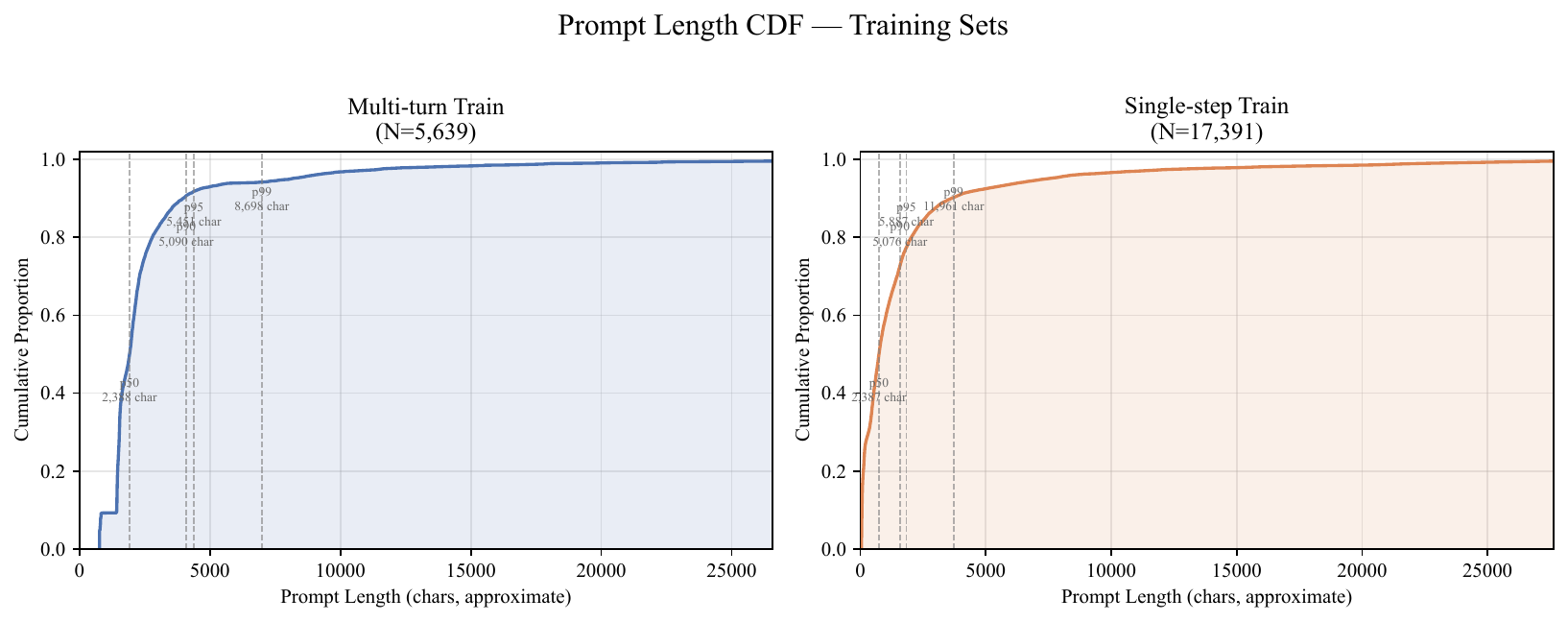}
\caption{Prompt-length CDF for the multi-turn and single-step training sets. Curves use character length for shape; dashed lines mark exact char percentiles.}
\label{fig:token_cdf}
\end{figure}

\subsection{Representative Data Sample}
\label{sec:data_sample}

Figure~\ref{fig:data-sample} shows a representative trajectory from the multi-turn training set, illustrating the Task Modification scenario. The user initially sets an alarm for 07:00 alongside enabling Bluetooth. In the subsequent turn, the user changes their mind and modifies the alarm time to 06:30. The agent must successfully update the arguments while strictly avoiding regression to the stale value (07:00), which requires precise intent tracking.

\begin{figure}[htb]
\centering

\begin{tcolorbox}[
  colback=gray!4!white,
  colframe=gray!85!black,
  title=\textbf{Candidate Tools (3 of 18 shown)},
  fonttitle=\bfseries,
  boxrule=0.8pt,
  top=2mm, bottom=2mm
]
\footnotesize
\textbf{remove\_schedule}: Delete a calendar event. Parameters: \textit{activity} (event name), \textit{date\_time} (date and time), \textit{position} (location), \textit{type\_of\_range} (scope: single/all/first N).\\[2pt]
\textbf{navigation\_tool}: Provide navigation from a departure to a destination. Parameters: \textit{departure\_location} (origin, required if specified), \textit{destination\_location} (target), \textit{type\_of\_action} (outbound/return).\\[2pt]
\textbf{transport\_service}: Ride-hailing service. Parameters: \textit{type\_of\_action} (start/cancel/query), \textit{department\_location} (pickup point).
\end{tcolorbox}

\begin{tcolorbox}[
  colback=gray!4!white,
  colframe=gray!85!black,
  title=\textbf{Turn 1: Initial Tool Call},
  fonttitle=\bfseries,
  boxrule=0.8pt,
  top=2mm, bottom=2mm
]
\footnotesize
\textbf{User}: Set an alarm for 7am tomorrow and turn on Bluetooth.

\textbf{Assistant}: I'll set up both for you.\\[2pt]
\texttt{<tool\_call>}\\
\texttt{\{"name": "alarm", "arguments": \{"time": "07:00", "repeat": "once"\}\}}\\
\texttt{\{"name": "bluetooth", "arguments": \{"action\_type": "open"\}\}}\\
\texttt{</tool\_call>}\\[2pt]
\textbf{Tool returns}: Alarm set for 07:00. Bluetooth enabled.
\end{tcolorbox}

\begin{tcolorbox}[
  colback=blue!2!white,
  colframe=blue!55!black,
  title=\textbf{Turn 2: Modification (parameter overwrite)},
  fonttitle=\bfseries,
  boxrule=0.8pt,
  top=2mm, bottom=2mm
]
\footnotesize
\textbf{User}: Actually, change the alarm to 6:30 instead.

\textbf{Assistant}: Got it, updating your alarm to 6:30.\\[2pt]
\texttt{<tool\_call>}\\
\texttt{\{"name": "alarm", "arguments": \{"time": "06:30", "repeat": "once"\}\}}\\
\texttt{</tool\_call>}\\[2pt]
\textbf{Tool return}: Alarm updated to 06:30.
\end{tcolorbox}

\begin{tcolorbox}[
  colback=teal!2!white,
  colframe=teal!80!black,
  coltitle=white,
  title=\textbf{Expected Output \& BeliefState},
  fonttitle=\bfseries,
  boxrule=0.8pt,
  top=2mm, bottom=2mm
]
\footnotesize
\textbf{Expected call}: \texttt{alarm(time="06:30", repeat="once")}\\[2pt]
\textbf{Stale value}: \texttt{time="07:00"} \quad (must not regress)\\[4pt]
\textbf{BeliefState}:\\
\quad $\mathcal{C}_{\text{slots}}$: alarm.time = 06:30 \texttt{[stale: 07:00]}\\
\quad $\mathcal{G}_{\text{current}}$: set alarm to 06:30\\
\quad $\mathcal{I}_{\text{signal}}$: modify\\[4pt]
\textbf{Scenario}: Task Modification \quad|\quad \textbf{Intent mode}: Modification\\
\textbf{Reward}: JGA hit $+$ SCRR stale penalty (if 07:00 is reused)
\end{tcolorbox}

\caption{A representative training trajectory from the DynamicIntent Dataset. Turn 1 shows an initial parallel tool call. Turn 2 modifies the alarm time from 07:00 to 06:30; the agent must invoke the correct tool with updated arguments and must not regress to the stale value 07:00. The BeliefState tracks the slot update with a stale flag, and the reward penalizes SCRR if the old value is reused.}
\label{fig:data-sample}
\end{figure}

\subsection{ID/OOD Partitioning and Test-Set Human Annotation}
\label{sec:annotation}

\noindent\textbf{ID/OOD partitioning.} The Intent Benchmark is split into ID and OOD along the tool-source axis. The ID set uses the same tool sources as training, while the OOD set uses a held-out, disjoint tool pool of 244 tools so that OOD scenarios exercise genuinely unseen schemas. Both splits cover the scenarios with approximately 100 dialogs each. The ID set contains 1,272 dialogs; the OOD set contains 1,278. The split is designed primarily to assess generalization to held-out tool schemas while maintaining comparable scenario coverage.

\noindent\textbf{Test-set human annotation protocol.} Every test dialog undergoes a structured annotation protocol to guarantee exact evaluation labels for the rule-based metrics. The protocol consists of three universal rules and nine scenario-specific checks. Each test instance was annotated by one annotator and independently verified by a second annotator. Disagreements were resolved by an adjudicator.

\textbf{Universal rules (Tier~1).} (1) \textbf{Parameter consistency}: every assistant tool call must conform to the declared tool schema---arguments must use the correct parameter names, types, and value ranges; undeclared keys are flagged as hallucinated. (2) \textbf{Closed-loop resolution}: each dialog must form a self-contained narrative in which the assistant's tool calls fully resolve the user's stated request; samples that terminate prematurely or diverge from the intended goal are flagged for revision or deletion. (3) \textbf{Revise-or-delete}: samples containing simple mistakes (e.g., a single misnamed parameter) are corrected in-place by the annotator; samples whose task execution substantially deviates from the expected outcome are discarded.

\textbf{Scenario-specific checks (Tier~2).} The two Simple Task scenarios are assessed using the universal rules in Tier~1. Nine scenario-specific verification categories cover the remaining benchmark scenarios. For \textbf{Parameter Clarification}, the annotator verifies that a mandatory parameter is genuinely omitted from the user query and that the assistant proactively asks for it. For \textbf{Irrelevant Reply} (both Simple Task and Complex Task variants), the annotator confirms that the user's reply introduces a genuinely new task and that the previously suspended old task is not required to be completed. For \textbf{Multi-turn Recursive Clarification}, both an intent-clarification and a parameter-clarification exchange must be present. For \textbf{Intent Clarification}, the user utterance must be deliberately ambiguous (neither chitchat nor a complete request), and the assistant's subsequent turn must clarify the intended meaning. For \textbf{Incomplete Prior Task} (both After-User and After-Tool variants), the annotator verifies that the interruption is actually injected at the correct position and that the newly introduced task is successfully resolved. For \textbf{Task Modification}, when the interruption occurs after a tool call, all non-modified arguments in the post-interruption call must remain identical to their pre-interruption values. For \textbf{Personalized Information Update}, the annotator checks that the profile update is logically coherent and that when the same tool is re-invoked after the update, at least one argument must differ due to the profile change; samples where the profile update leaves all arguments unchanged are either manually modified to link the updated field to a tool parameter or deleted. For \textbf{Parallel Multi-task Execution}, the annotator verifies that the combined user request contains two independent subtasks and that the trajectory resolves both without omitting or conflating either goal. Tool calls and argument values must correspond to the correct subtask, with no cross-task slot contamination or unintended dependency between the subtasks.
Any tool dependencies within each subtask must remain valid. For \textbf{Dependency Chain}, the annotator confirms that a legitimate sequential dependence exists between two or more tool calls (i.e., the output of tool A feeds the input of tool B) and that the chain completes correctly.

The test set totals 2,550 annotated dialogs (1,272 ID, 1,278 OOD). Approximately 18.7\% of auto-generated samples are revised in-place and 4.2\% are discarded during annotation, ensuring that the final evaluation labels meet the precision requirements of the rule-based metrics. Annotation was carried out by an internal annotator team at the authors' institution: no external participants were recruited, so no recruitment or compensation procedure applies, and the task involved no personal data about the annotators.

\subsection{Multi-dimensional Intent Metrics}
\label{sec:metrics}
\begin{figure}[htb]
\centering
\includegraphics[width=0.80\linewidth]{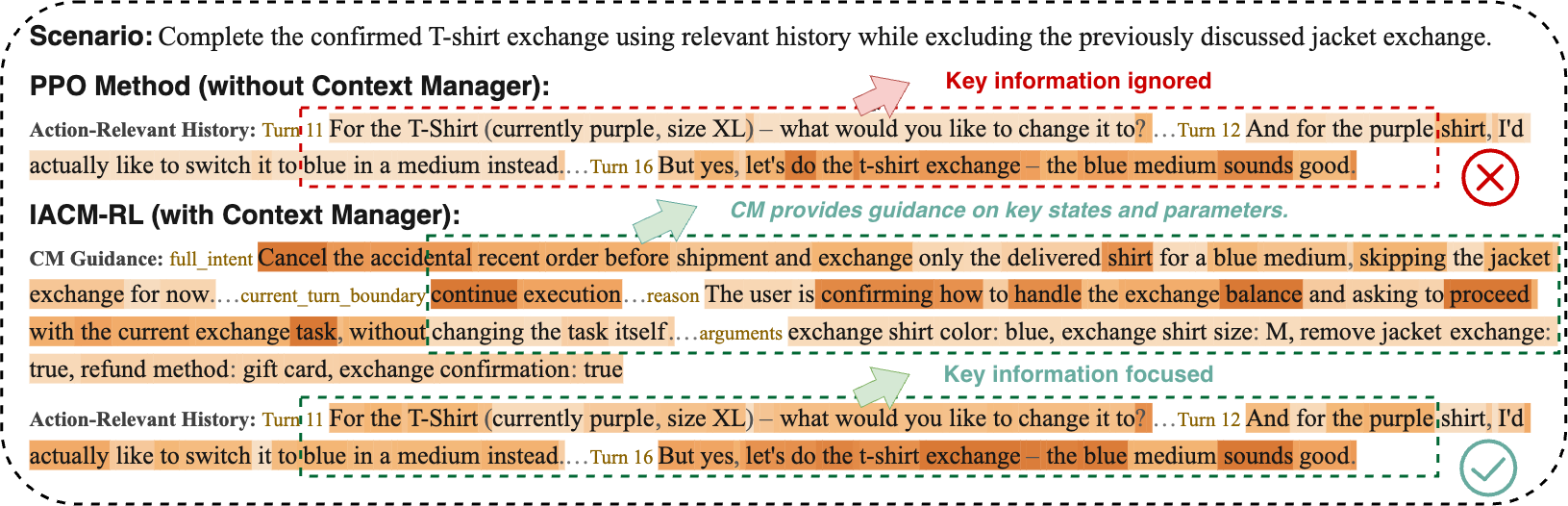}
\caption{Attention-level analysis of a failure case under dynamic intent fluctuations. (Top) Conventional paradigms let obsolete constraints and injected noise during drift turns dilute token-level attention across verbose histories, losing focus on critical keywords. (Bottom) IACM-RL deploys a Context Manager that structurally isolates overwritten parameters and refocuses attention on valid state boundaries and current goals. Darker colors indicate higher attention.}
\label{fig:motivation}
\end{figure}

Single-turn accuracy fails to capture the cascading failures of long-horizon intent fluctuation, so we design a five-metric matrix diagnosing dynamic fault-tolerance and structural compliance. The five metrics differ in statistical unit, so each definition below fixes its own unit and aggregation.

\noindent\textbf{Joint Goal Accuracy (JGA)} grades every turn $t$ by comparing the predicted call multiset $\hat{\mathcal{A}}_{t}$ with the gold multiset $\mathcal{A}^{*}_{t}$, where $\equiv$ matches tool names and multiplicities only and $=$ additionally requires identical argument values:
\begin{equation}
\small
\begin{gathered}
\text{JGA} = \frac{100}{2\,|\mathcal{T}_{\text{eval}}|} \sum_{t \in \mathcal{T}_{\text{eval}}} q_t,\\
q_t = \mathbf{I}\!\left(\hat{\mathcal{A}}_{t} \equiv \mathcal{A}^{*}_{t}\right) + \mathbf{I}\!\left(\hat{\mathcal{A}}_{t} = \mathcal{A}^{*}_{t}\right)
\end{gathered}
\label{eq:jga}
\end{equation}

\noindent\textbf{Stale Context Residual Rate (SCRR)} measures how often a modification turn regresses to a value that the modification overwrote, where $\mathcal{S}_t$ collects the gold parameter pairs dropped at turn $t$ and $\hat{\mathcal{P}}_t$ the pairs the agent predicts:
\begin{equation}\small
    \text{SCRR} = \frac{100}{|\mathcal{T}_{\text{mod}}|} \sum_{t \in \mathcal{T}_{\text{mod}}} \mathbf{I}\!\left( \hat{\mathcal{P}}_t \cap \mathcal{S}_t \neq \varnothing \right)
    \label{eq:scrr}
\end{equation}

\noindent\textbf{Redundant and Infinite-loop Rate (RIR)} counts problem calls over the $N$ issued calls, where $d_j$ flags a call repeating an earlier call of the same dialogue and $z_j$ flags a call whose tool is absent from the available tool list:
\begin{equation}\small
    \text{RIR} = \frac{100}{N} \sum_{j=1}^{N} \left[\, \mathbf{I}(d_j) + \mathbf{I}(z_j) \right]
    \label{eq:rir}
\end{equation}
A call may be both, so RIR is an intensity of problem calls rather than a bounded proportion.

\noindent\textbf{Dependency Tool Chain Rate (DTCR)} scores chain completion per dialogue, where an LLM judge compares the gold and predicted call chains and returns a rubric score $q_d \in \{0,1,2\}$ ($2$: order and data flow complete; $1$: partially correct; $0$: a critical call missing or wrong):
\begin{equation}\small
    \text{DTCR} = \frac{100}{2\,|\mathcal{D}_{\text{chain}}|} \sum_{d \in \mathcal{D}_{\text{chain}}} q_d
    \label{eq:dtcr}
\end{equation}

\noindent\textbf{Intent Switch Success Rate (ISSR)} scores pivoting on interruption turns, where an LLM judge compares the agent's calls and response at turn $t$ against the previous turn's gold and returns $s_t \in [-1,1]$:
\begin{equation}\small
    \text{ISSR} = \frac{100}{|\mathcal{T}_{\text{int}}|} \sum_{t \in \mathcal{T}_{\text{int}}} \frac{s_t + 1}{2}
    \label{eq:issr}
\end{equation}
ISSR is a normalised switch-quality score: $50$ marks the scale midpoint, not half of the interruptions succeeding.

\section{Methodology}

\subsection{Problem Formulation and Motivation}
We formulate multi-turn tool invocation under dynamic intent noise as a POMDP $\mathcal{M} = \langle \mathcal{S}, \mathcal{A}, \mathcal{T}, \mathcal{R},\allowbreak \Omega, \mathcal{O}, \gamma \rangle$, where $\mathcal{S}$ is the latent intent state, $\mathcal{A}$ the action space (linguistic responses and tool calls), $\mathcal{T}$ the state transition, $\mathcal{R}$ the reward, $\Omega$ the observation space, $\mathcal{O}$ the observation function, and $\gamma$ the discount factor. The latent intent $\mathcal{S}$ is only partially observed through noisy observations (mid-task modifications, irrelevant chit-chat, ambiguous constraints). Conventional imitation learning implicitly scans the raw history $h_{t-1}$ to optimize $\pi(a_t \mid h_{t-1}, o_t)$, bundling state tracking and action generation into a single black box.

This work is motivated by an attention-level diagnosis of some typical failures exhibited by conventional policies learning under dynamic intent fluctuations (Figure~\ref{fig:motivation}). During drift turns, such as mid-task modifications or interruptions, the accumulation of obsolete constraints and injected noise dilutes the model's token-level attention across the verbose dialogue history. Consequently, critical tokens representing the updated goal or the latest parameter changes receive low attention weights, causing the agent to overlook key information. This attention dilution can undermine the data dependencies required for multi-tool execution, increasing the risk of intent deviation and repetitive API calls.

\subsection{Self-Generated Context Manager via BeliefState}
To systematically neutralize the error amplification identified above, we introduce an autoregressive Self-Generated Context Manager. During training, at each turn the policy first decodes a compact state block from the current observation and history, then conditions the downstream tool action on this block together with the raw history. The block is governed by an explicit BeliefState with five core components (instantiated as nine XML sub-blocks, Figure~\ref{fig:cm-structure}):
\begin{equation}\small
    \mathbf{b}_t = \Big( \mathcal{C}_{\text{slots}}, \mathcal{G}_{\text{current}}, \mathcal{A}_{\text{last}}, \mathcal{Q}_{\text{pending}}, \mathcal{I}_{\text{signal}} \Big)
    \label{eq:beliefstate}
\end{equation}
where $\mathbf{b}_t$ is the BeliefState at turn $t$, $\mathcal{G}_{\text{current}}$ captures the core objective, $\mathcal{I}_{\text{signal}}$ the intent deviation category such as clarify, modify, or interrupt, $\mathcal{A}_{\text{last}}$ the most recent tool execution for loop suppression, $\mathcal{Q}_{\text{pending}}$ the missing information that requires clarification, and $\mathcal{C}_{\text{slots}}$ the verified parameters. Crucially, when a parameter is overwritten, the old value is not deleted but kept and tagged with a structural stale flag that marks it as no longer usable.

\begin{figure}[htb]
\centering
\begin{tcolorbox}[
    colback=gray!4!white,
    colframe=gray!50!black,
    coltitle=white,
    title=\textbf{Context Manager Block Structure},
    fonttitle=\normalsize,
    boxrule=0.5pt,
    arc=2mm,
    top=2mm, bottom=2mm, left=3mm, right=3mm
]
\footnotesize
\begin{tabular}{@{}p{15.4cm}@{}}
\hspace*{-0.3cm}\texttt{<Context\_Manager>}\\[2pt]
<Current\_User> intent \& tool-call flag </Current\_User>\\[1pt]
<Overall\_User\_Task> goal \& deferred intents </Overall\_User\_Task>\\[1pt]
<History\_Relation> inherit / override / delete slots </History\_Relation>\\[1pt]
<Resolved\_References> resolved entity references </Resolved\_References>\\[1pt]
<Active\_Objects> last execution \& active objects </Active\_Objects>\\[1pt]
<Pending\_Or\_Interrupted\_Tasks> blocked \& interrupted tasks </Pending\_Or\_Interrupted\_Tasks>\\[1pt]
<Task\_Graph\_For\_Current\_Turn> slots \& tool plan </Task\_Graph\_For\_Current\_Turn>\\[1pt]
<Deferred\_Tasks\_After\_Current\_Turn> postponed sub-goals </Deferred\_Tasks\_After\_Current\_Turn>\\[1pt]
<Tool\_Decision\_Guide> call / ask \& unresolved slots </Tool\_Decision\_Guide>\\[3pt]
\hspace*{-0.3cm}\texttt{</Context\_Manager>}\\
\end{tabular}
\end{tcolorbox}
\caption{The nine sub-blocks of the self-generated Context Manager. The CM update timing is reported in Section~\ref{sec:app_cm_timing}.}
\label{fig:cm-structure}
\end{figure}

\subsection{Hierarchical Intent-Driven Reward System}
To provide dense, diagnostically meaningful reinforcement, we decompose the reward into three layers mirroring the cognitive, behavioral, and outcome aspects of tool calling:
\begin{equation}\small
    R_{\text{total}} = \alpha\, R_{\text{belief}} + \beta\, R_{\text{action}} + \gamma\, R_{\text{outcome}} + R_{\text{format}}
\end{equation}
where $\alpha,\beta,\gamma$ are the layer weights and $R_{\text{format}}$ is a small penalty on malformed responses such as empty outputs or residual tool-call tags. Each layer targets the failure modes measured by the metrics of Section~3.2 and supervises a subset of the BeliefState fields: $R_{\text{belief}}$ combines JGA with stale and hallucination penalties to supervise $\mathcal{C}_{\text{slots}}$; $R_{\text{action}}$ incorporates loop and invalid-call penalties as well as LLM-evaluated Active Clarification Reward (ACR) to supervise $\mathcal{A}_{\text{last}}$ and $\mathcal{Q}_{\text{pending}}$; and $R_{\text{outcome}}$ combines DTCR and ISSR to supervise $\mathcal{G}_{\text{current}}$ and $\mathcal{I}_{\text{signal}}$. This yields a direct correspondence between reward layers and BeliefState fields, turning the opaque trajectory reward into field-level credit assignment. Layer weights and all signal coefficients are reported in Section~\ref{sec:reward_details} and Section~\ref{sec:app_exp}.

\subsection{Intent-Driven RL Optimization via CM Auxiliary Losses}
We optimize $\pi_\theta$ via PPO with GAE. Since the CM is self-generated, we drop the cross-entropy imitation loss and introduce two auxiliary losses:
\begin{equation}\small
    \mathcal{L}_{\text{total}} = \mathcal{L}_{\text{PG}} + w_{\text{ext}}\mathcal{L}_{\text{ext}} + w_{\text{dist}}\mathcal{L}_{\text{dist}}
\end{equation}
where $\mathcal{L}_{\text{PG}}$ is the standard PPO policy-gradient loss, $w_{\text{ext}},w_{\text{dist}}$ are balancing weights, and $\tau$ denotes a sampled trajectory.

\noindent\textbf{CM Extraction.} The core loss making the CM self-improving: it directly optimizes the CM block tokens with the trajectory advantage, closing the loop from state quality to reward:
\begin{equation}\small
    \mathcal{L}_{\text{ext}} = -\mathbf{E}_{\tau}\!\left[\frac{\sum_{i\in\mathbf{b}_t}\log\pi_\theta(b_{t,i})\,m_i\,\hat{A}_\tau}{\sum_{i\in\mathbf{b}_t}m_i}\right]
\end{equation}
where $b_{t,i}$ is the $i$-th token of the CM block $\mathbf{b}_t$ at turn $t$, $m_i$ is the binary CM token mask indicating which tokens belong to the CM block, $\hat{A}_\tau$ is the detached trajectory advantage aggregated from token-level advantages, and the sum is over all tokens in the CM block.

\noindent\textbf{State Distillation.} Distills the CM-conditioned teacher into a CM-free student via reverse KL, so state tracking is internalized into the weights and survives without the explicit block:
\begin{equation}\small
    \mathcal{L}_{\text{dist}} = \mathbf{E}_{\tau}\!\left[\,D_{\text{KL}}\!\left(\pi^{S}_\theta\;\big\|\;\pi^{T}_\theta{}_{\text{sg}}\right)\right]
\end{equation}
where $\pi^{T}_\theta \equiv \pi_\theta(\cdot\mid\mathbf{b}_t,h_{t-1},o_t)$ is the teacher policy conditioned on the CM block, $\pi^{S}_\theta \equiv \pi_\theta(\cdot\mid h_{t-1},o_t)$ is the student policy without the CM block ($h_{t-1}$ is the dialogue history up to turn $t$ and $o_t$ the current observation), and the subscript $\text{sg}$ detaches the teacher distribution so only the student receives gradients.

The auxiliary objectives operate on complementary targets: CM extraction reinforces generated context tokens using trajectory advantages, and state distillation matches action distributions across CM-conditioned and CM-free inputs.

\subsection{Training Algorithm}
\label{sec:app_algorithm}

The complete two-stage training procedure is detailed in Algorithm~\ref{alg:iacmrl}. Both stages use the same three-component CM pipeline: a BeliefState updater $\mathcal{U}$ that maintains the structured state from observable messages, a deterministic XML template that renders a target CM from that state, and a self-drafted CM block that the policy produces by decoding from $(\mathbf{b}_t, h_t)$. Stage~1 establishes tool-selection proficiency on single-turn slices; Stage~2 initializes from Stage~1 and introduces the hierarchical reward and the two auxiliary losses. Due to space constraints, all mathematical expressions within the pseudocode are compactly formatted and may differ slightly from their expanded forms presented in the main text.

\begin{algorithm}[t]
\caption{Graph-Constrained Trajectory Splicing}
\label{alg:splice}
\begin{algorithmic}[1]
\REQUIRE Base trajectory $A$; candidate pool $\mathcal{B}$; dependency graph $G$; insertion position $p$.
\STATE $\mathcal{T}_A \leftarrow$ tool names invoked in $A$
\FOR{each candidate $B \in \mathcal{B}$ (shuffled)}
    \STATE $\mathcal{T}_B \leftarrow$ tool names invoked in $B$
    \IF{$\mathcal{T}_A \cap \mathcal{T}_B \neq \emptyset$}
        \STATE \textbf{continue} \COMMENT{name overlap: not orthogonal}
    \ENDIF
    \STATE $\textit{conflict} \leftarrow \textbf{false}$
    \FOR{each $(t_a, t_b) \in \mathcal{T}_A \times \mathcal{T}_B$}
        \STATE $u_a \leftarrow \textsc{NameToUuid}(t_a)$; \quad $u_b \leftarrow \textsc{NameToUuid}(t_b)$
        \IF{$(u_a, u_b) \in G$ \textbf{or} $(u_b, u_a) \in G$}
            \STATE $\textit{conflict} \leftarrow \textbf{true}$; \textbf{break}
        \ENDIF
    \ENDFOR
    \IF{$\neg\,\textit{conflict}$}
        \STATE Re-index the round and tool-call identifiers of $B$ from $\textsc{MaxRound}(A)+1$.
        \STATE Insert $B$ into $A$ at position $p$; merge and de-duplicate the tool pools.
        \RETURN spliced trajectory $A'$
    \ENDIF
\ENDFOR
\RETURN \textsc{Failure} \COMMENT{no orthogonal candidate; $A$ discarded}
\end{algorithmic}
\end{algorithm}

\subsection{Reward Function Details}
\label{sec:reward_details}

\noindent\textbf{Scope.} The metrics of Section~\ref{sec:metrics} measure held-out performance, whereas the training reward is a separate and  denser surrogate targeting the same failure modes: it is free to use different matching rules, aggregation units and scales, and the two are not numerically comparable. This appendix specifies the surrogate only. Let $\hat{\mathcal{A}}_t$ be the set of tool calls predicted by the agent at turn $t$, each comprising a function name $f$ and arguments $\{k_i:v_i\}$; let $\mathcal{A}^*_t$ be the gold calls from the data pipeline; and let $\mathcal{T}_{\text{mod}}, \mathcal{T}_{\text{int}}, \mathcal{T}_{\text{chain}}$ denote the turns labelled Modification, Interruption and Chaining respectively.

\noindent\textbf{Joint Goal Accuracy (JGA).} In place of the graded rubric of Eq.~\eqref{eq:jga}, the reward uses a call-level hit rate: a predicted call hits when its tool name and argument keys match a gold call, each gold entry credited at most once, giving $\frac{n_{\text{hit}}}{n_{\text{calls}}} \times 2.5$ over the calls actually issued (the single-step variant uses $\frac{2 \cdot n_{\text{hit}}}{n_{\text{calls}}+1}$). When the executor reports no per-call outcome, the hit count is recomputed from the predicted calls against the same gold entries, so both routes share one scale.

\noindent\textbf{Stale Context Residual Rate (SCRR).} The reward flags regression to a superseded value rather than generic argument mismatch: reading each (tool, argument key) gold sequence in order, a call whose value sits earlier than one the trajectory has already reached counts as one violation at $-0.5$. Trajectory order is what keeps the rule from penalising a policy that correctly skips an obsolete intent. An argument value absent from the observable context (all user turns together with the tool returns) costs $-0.8$. SCRR is disabled in Stage~1 because a single step cannot exhibit cross-turn overwriting.

\noindent\textbf{Redundant and Infinite-Loop Rate (RIR).} The reward instantiates the failure modes of Eq.~\eqref{eq:rir} through separate penalties rather than a single ratio. Two adjacent calls sharing a tool name and identical arguments after key-order normalisation short-circuit the trajectory to $-2.0$, with the diagnostic counters still populated. An out-of-pool tool name costs $-1.0$ each in Stage~2; the single-step variant applies no invalid-call term, as it has no schema check. In Stage~2 only, duplicates beyond twice the required call count cost $-0.5$ per excess call, and more than three calls with a hit rate below $0.3$ costs $-0.5$. Trajectory progress adds $+0.3$ per distinct (tool, arguments) pair beyond the first, capped at $+0.9$. These magnitudes balance the hierarchical sum and carry no direct correspondence to the evaluation ratio.

\begin{algorithm}[t]
  \caption{IACM-RL Two-Stage Training Algorithm}
  \label{alg:iacmrl}
  \begin{algorithmic}[1]
  \REQUIRE Dataset $\mathcal{D}$, SFT checkpoint $\theta_0$, weights $w_{\text{ext}}, w_{\text{dist}}$ and $\alpha, \beta, \gamma$.
  \ENSURE Optimized policy $\pi_\theta$.
  \STATE \textbf{Stage 1: single-step training with self-generated CM.} Enable $\mathcal{R}_{\text{single}}$ and sample single-turn slices from $\mathcal{D}_{\text{step}}$.
  \FOR{each iteration}
      \STATE \textbf{Rollout:} ingest the user message and update the BeliefState $\mathbf{b}$; decode the CM block from the history; inject the CM into the system prompt and generate one response $a$ (tool call or text); if $a$ is a tool call, execute it via MockBackend and record the result in $\mathbf{b}$.
      \STATE Compute the reward $\mathcal{R}_{\text{single}}$ (JGA $+$ ACR, no CM losses) and the GAE advantages $\hat{A}_t$.
      \STATE $\mathcal{L} \leftarrow \mathcal{L}_{\text{PG}} = -\mathbb{E}_t\!\left[\min\!\left(\rho_t \hat{A}_t,\ \operatorname{clip}\!\left(\rho_t, 1-\epsilon, 1+\epsilon\right) \hat{A}_t\right)\right]$
      \STATE Update $\theta$ via $\nabla_\theta \mathcal{L}$.
  \ENDFOR
  \STATE Save the Stage-1 checkpoint $\theta_1$.
  \STATE \textbf{Stage 2: multi-turn training with self-generated CM.} Load $\theta_1$; enable $\mathcal{R}_{\text{total}}$ and the auxiliary losses.
  \FOR{each iteration}
      \STATE Sample multi-turn trajectories from $\mathcal{D}_{\text{multi}}$.
      \FOR{each turn $t$}
          \STATE \textbf{Rollout:} ingest the user message and update $\mathbf{b}_t$ (slots, goal, intent, last action); decode the CM block from the history; inject it into the system prompt and generate the response $a_t$ (tool call or text); if $a_t$ is a tool call, execute it via MockBackend and record the result in $\mathbf{b}_t$.
      \ENDFOR
      \STATE \textbf{Hierarchical reward:} $R_{\text{belief}}$ (JGA hits $-$ SCRR stale $-$ hallucination), $R_{\text{action}}$ (loop/invalid/precision penalties $+$ trajectory progress $+$ ACR clarification), $R_{\text{outcome}}$ (DTCR chain completion $+$ ISSR switch success).
      \STATE $R_{\text{total}} = \alpha R_{\text{belief}} + \beta R_{\text{action}} + \gamma R_{\text{outcome}} + R_{\text{format}}$; collect the trajectory $\tau = \{(s_t, a_t, r_t)\}$.
      \STATE Compute the GAE advantages $\hat{A}_t$ and the trajectory advantage $\hat{A}_\tau$.
      \STATE \textbf{Auxiliary losses:} $\mathcal{L}_{\text{ext}} = -\mathbb{E}_\tau\!\left[\frac{\sum_{i} \log \pi_\theta(b_{t,i})\, m_i\, \hat{A}_\tau}{\sum_i m_i}\right]$, $\mathcal{L}_{\text{dist}} = \mathbb{E}_\tau\!\left[D_{\text{KL}}\!\left(\pi^{S}_\theta \,\big\|\, \pi^{T}_\theta{}_{\text{sg}}\right)\right]$.
      \STATE $\mathcal{L}_{\text{total}} = \mathcal{L}_{\text{PG}} + w_{\text{ext}}\mathcal{L}_{\text{ext}} + w_{\text{dist}}\mathcal{L}_{\text{dist}}$
      \STATE Update $\theta$ via $\nabla_\theta \mathcal{L}_{\text{total}}$.
  \ENDFOR
  \RETURN $\pi_\theta$.
  \end{algorithmic}
\end{algorithm}

\noindent\textbf{Action Correction Rate (ACR).} When the expected response is a natural-language clarification rather than a tool call, an LLM judge (GPT-5.4) scores the agent's response quality along relevance, specificity, and completeness, producing $s_{\text{LLM}}\in[-1,1]$. In the single-step reward, the raw score directly feeds the reward (halved for pure-query turns). In the multi-turn reward (Stage~2), the ACR reward is $+1.0\times s_{\text{LLM}}$ when $s_{\text{LLM}}>0$, active only on Clarification-labeled turns. ACR is a reward-only signal and is not part of the evaluation matrix of Section~\ref{sec:metrics}: it applies when no tool call is issued and the question type is language or query.

\noindent\textbf{Intent Switch Success Rate (ISSR).} On $\mathcal{T}_{\text{int}}$, an LLM judge compares the current turn's tool selection against a gold post-interruption reference, producing $s_{\text{LM}}\in[-1,1]$. Unlike the evaluation indicator (Eq.~\eqref{eq:issr}), the training-time reward is score-weighted so that stronger pivots receive proportionally larger positive signals:
\begin{equation}
    R_{\text{ISSR}} = \lambda_{\text{switch}}\,s_{\text{LM}}\;\mathbf{I}(s_{\text{LM}}>\tau_{\text{ISSR}}),
\end{equation}
with $\lambda_{\text{switch}}\!=\!2.0$ and $\tau_{\text{ISSR}}\!=\!0.3$ in Stage~2. The threshold gate is a training-time choice: the reward is zero below $\tau_{\text{ISSR}}$ and scaled by the judge score above it, whereas the evaluation metric of Eq.~\eqref{eq:issr} averages $s_{\text{LM}}$ over interruption turns without a threshold.

\noindent\textbf{Dependency Tool Chain Rate (DTCR).} On $\mathcal{T}_{\text{chain}}$ where interdependent calls are required, the chain completion ratio $c$ is the fraction of $\mathcal{A}^*_t$ correctly called (JGA hits). Stage~2 uses a graded scheme:
\begin{equation}
    R_{\text{chain}}(c)\!=\!
    \begin{cases}
        +4.0 & c\!=\!100\%,\\
        +2.5 & 80\%\!\leq\! c\!<\!100\%,\\
        +1.0 & 50\%\!\leq\! c\!<\!80\%
    \end{cases}
\end{equation}

\begin{table}[htb]
\centering
\small
\begin{tabular}{lcccp{5.4cm}}
\toprule
\textbf{Signal} & \textbf{Layer} & \textbf{Single-step } & \textbf{Multi-turn } & \textbf{Value / Rule} \\
\midrule
JGA (argument hit) & Belief & \checkmark & \checkmark & Single: $\frac{2 \cdot n_{\text{hit}}}{n_{\text{calls}}+1}$ / Multi: $\frac{n_{\text{hit}}}{n_{\text{calls}}} \times 2.5$ \\
SCRR stale param & Belief & -- & \checkmark & $-0.5$ per regression to a superseded value \\
SCRR hallucinated param & Belief & -- & \checkmark & $-0.8$ per value absent from user turns and tool returns \\
RIR infinite loop & Action & -- & \checkmark & $-2.0$, early return \\
RIR invalid call & Action & -- & \checkmark & $-1.0$ per out-of-pool tool \\
Trajectory progress & Action & -- & \checkmark & $+0.3$ per distinct call beyond the first (cap: $0.9$) \\
Precision penalties & Action & -- & \checkmark & Redundant: $-0.5$/call; Low-precision: $-0.5$ \\
ACR clarification & Action & \checkmark & \checkmark & LLM judge $\in[-1.0,1.0]$ (Multi: $+1.0 \times \text{score}$ if $>0$) \\
DTCR chain completion & Outcome & -- & \checkmark & $+4.0$ ($100\%$) / $+2.5$ ($\geq 80\%$) / $+1.0$ ($\geq 50\%$) \\
ISSR intent switch & Outcome & -- & \checkmark & LLM judge, $+2.0 \times \text{score}$ if score $>0.3$ \\
\bottomrule
\end{tabular}
\caption{Reward signal comparison between the single-step and multi-turn variants. \checkmark: active; --: not applicable.}
\label{tab:reward_compare}
\end{table}

The two training stages instantiate two reward variants that share the same detector primitives but differ in scope. Both parse the model response with a multi-call extractor, then branch on whether any tool call was produced.

\noindent\textbf{Single-step reward}. Designed for one-shot calls where the expected tool calls form a flat name-to-arguments map with no multi-tool sequence. It operates as four mutually exclusive branches keyed on question type:
\begin{enumerate}[leftmargin=*,itemsep=1pt,topsep=2pt]
    \item \emph{Tool call issued but question type is language}: return $-1.0$, the agent should have clarified.
    \item \emph{Tool call issued and question type is not language}: compute JGA as $\frac{2 \cdot n_{\text{hit}}}{n_{\text{calls}}+1}$, with loose matching for in-domain data and strict matching for BFCL.
    \item \emph{No tool call and question type is language}: invoke the ACR rubric on the natural-language response, yielding a score in $[-1,1]$.
    \item \emph{No tool call and question type is query}: invoke the same ACR rubric but halve the score to down-weight ungrounded answers.
\end{enumerate}
Malformed responses receive $-0.3$ to $-0.5$ format penalties. No loop, chain, or switch detection is applied, since a single step cannot exhibit them.

\noindent\textbf{Multi-turn reward}. Covers full trajectories and aggregates the three hierarchical layers described in the main text, with $\alpha{=}1.0$, $\beta{=}1.0$, $\gamma{=}0.5$, and final score clamping to $[-4.0, 8.0]$. Each detector is gated by the scenario label to bound LLM judging to at most 2 calls per sample. A consecutive-identical-call loop short-circuits to $-2.0$ and returns early.
\begin{enumerate}[leftmargin=*,itemsep=1pt,topsep=2pt]
    \item \textbf{Cognitive Reward Layer ($R_{\text{belief}}$).} Joint Goal Accuracy (JGA) grants positive signals based on the hit rate, $(n_{\text{hit}}/n_{\text{calls}}) \times 2.5$, matched greedily against a per-tool gold queue. Rule-based penalties counter memory failures: regressing to a superseded parameter value costs $-0.5$ per violation, and fabricating an argument value absent from the user turns and tool returns costs $-0.8$. $R_{\text{belief}}$ directly supervises the Confirmed Slots field $\mathcal{C}_{\text{slots}}$ of the BeliefState.

    \item \textbf{Behavioral Reward Layer ($R_{\text{action}}$).} Action rewards consist of several precision heuristics: a call to a tool outside the active schema costs $-1.0$; redundant calls beyond the required limit cost $-0.5$ each; and more than three calls at a hit rate below $0.3$ incurs a precision penalty of $-0.5$. Conversely, \emph{Trajectory Progress} rewards $+0.3$ per distinct call beyond the first (capped at $+0.9$). For clarification, when mandatory parameters are missing, an LLM judge evaluates the agent's proactive questions (ACR), granting $+1.0 \times \text{score}$ if the score exceeds $0.0$. $R_{\text{action}}$ supervises the Last Action $\mathcal{A}_{\text{last}}$ and Pending Questions $\mathcal{Q}_{\text{pending}}$ fields.

    \item \textbf{Outcome Reward Layer ($R_{\text{outcome}}$).} Dependency Tool Chain Rate (DTCR) provides graded reinforcement based on chain completion rates: $100\%$ completion yields $+4.0$, $\geq 80\%$ yields $+2.5$, and $\geq 50\%$ yields $+1.0$. Intent Switch Success Rate (ISSR) grants $+2.0 \times \text{score}$ when the agent successfully pivots to an interruption goal, gated by an LLM judge threshold of $0.3$. $R_{\text{outcome}}$ supervises the Current Goal $\mathcal{G}_{\text{current}}$ and Intent Signal $\mathcal{I}_{\text{signal}}$ fields.
\end{enumerate}

The three reward layers map one-to-one onto BeliefState fields, turning the opaque trajectory reward into field-level credit assignment. LLM-judged metrics (ACR, ISSR) use \texttt{gpt-5.4} with a prompt template adapted from the evaluation rubric.

\noindent\textbf{Signal comparison.} Table~\ref{tab:reward_compare} contrasts the two variants, detailing the precise penalty rules and grading thresholds applied in our optimization pipeline.

\subsection{CM Update Timing}
\label{sec:app_cm_timing}
The BeliefState is updated at well-defined points, and the granularity differs by stage. In single-step training, the CM is initialized and rendered once before generation, then the tool result is recorded once after generation. In multi-turn training, the state is updated at every interaction boundary: each user message triggers intent detection, slot and stale-flag updates, and goal revision; each tool execution triggers a last-action and active-object update, followed by a fresh render that re-injects the updated CM into the system prompt for the next assistant turn.

\section{Experiments}

\begin{table}[htb]
\centering
\footnotesize
\begin{tabular}{lcccccccccc}
\toprule
\multirow{2}{*}{\textbf{Method}} & \multicolumn{3}{c}{\textbf{Intent ID}} & \multicolumn{3}{c}{\textbf{Intent OOD}} & \multicolumn{2}{c}{\textbf{$\mathrm{\tau}^2$-Bench}} & \multirow{2}{*}{\textbf{BFCL}} & \multirow{2}{*}{\textbf{Avg}} \\
\cmidrule(lr){2-4} \cmidrule(lr){5-7} \cmidrule(lr){8-9}
 & Cog & Beh & Out & Cog & Beh & Out & Air & Ret & & \\
\midrule
Base         & 34.8 & 99.4 & 84.1 & 33.0 & 99.4 & 77.7 & 28.0 & 31.6 & 63.1 & 61.2 \\
PPO-noCM      & 34.8 & \textbf{99.9} & 85.6 & 32.4 & 99.6 & 81.0 & 32.0 & 26.3 & 62.2 & 61.5 \\
FIFO-k       & 33.3 & 99.5 & 84.8 & 31.4 & 99.4 & 82.5 & 30.0 & 28.9 & 58.9 & 61.0 \\
Prompt-Comp  & 32.2 & 99.8 & 86.3 & 34.0 & 98.3 & 81.5 & 30.0 & 21.9 & 61.6 & 60.6 \\
ACON         & 33.7 & 99.4 & 83.5 & 34.7 & 99.2 & 81.7 & 26.0 & 35.1 & 61.4 & 61.6 \\
LLMLingua    & 29.6 & \textbf{99.9} & 84.7 & 31.2 & \textbf{99.8} & 79.5 & 28.0 & \textbf{38.6} & 62.2 & 61.5 \\
MEM1         & 36.4 & \textbf{99.9} & \textbf{87.3} & 34.7 & 99.2 & 82.7 & \textbf{38.0} & 23.7 & 61.9 & 62.6 \\
RL-STA       & 36.4 & 99.7 & 86.6 & 33.8 & 99.3 & \textbf{83.0} & 28.0 & 27.4 & 63.2 & 62.0 \\
\rowcolor{gray!20}\textbf{IACM-RL (Ours)} & \textbf{36.5} & \textbf{99.9} & \textbf{87.3} & \textbf{36.3} & 99.6 & 81.2 & \textbf{38.0} & 34.2 & \textbf{63.3} & \textbf{64.0} \\
\bottomrule
\end{tabular}
\captionof{table}{Main results on the Intent Benchmark, $\mathrm{\tau}^2$-Bench, and BFCL-V3.
IACM-RL achieves the highest overall average and strong tool-calling performance on complex intent-fluctuation scenarios.
Compression baselines lag on Cognitive due to history truncation.
On $\mathrm{\tau}^2$-Bench, IACM-RL leads on airline and stays competitive on retail.
Cognitive = (JGA$-$SCRR)/2, Behavioral = 100$-$RIR, Outcome = (ISSR+DTCR)/2; RIR is a problem-call intensity (Eq.~\eqref{eq:rir}) and stays below 2\% in all reported settings.
Best in \textbf{bold}.}
\label{tab:main}
\vspace{8pt}
\begin{tabular}{lcccccccccc}
\toprule
\multirow{2}{*}{\textbf{Variant}} & \multicolumn{3}{c}{\textbf{Intent ID}} & \multicolumn{3}{c}{\textbf{Intent OOD}} & \multicolumn{2}{c}{\textbf{$\mathrm{\tau}^2$-Bench}} & \multirow{2}{*}{\textbf{BFCL}} & \multirow{2}{*}{\textbf{Avg}} \\
\cmidrule(lr){2-4} \cmidrule(lr){5-7} \cmidrule(lr){8-9}
 & Cog & Beh & Out & Cog & Beh & Out & Air & Ret & & \\
\midrule
\rowcolor{gray!20}\textbf{Full (OURS)} & 36.5 & 99.9 & \textbf{87.3} & \textbf{36.3} & 99.6 & 81.2 & \textbf{38.0} & \textbf{34.2} & \textbf{63.3} & \textbf{64.0} \\
$-\mathcal{L}_{\text{ext}}$   & \textbf{37.0} & 99.9 & 85.5 & 32.7 & 99.0 & 82.3 & 30.0 & 30.7 & 62.9 & 62.2 \\
$-\mathcal{L}_{\text{dist}}$  & 34.6 & \textbf{100.0} & 85.2 & 31.4 & \textbf{100.0} & 81.0 & 36.0 & 30.9 & 61.1 & 62.2 \\
$-$all aux & 35.8 & 99.8 & 86.7 & 33.7 & 99.0 & \textbf{82.6} & 28.0 & 25.6 & 62.1 & 61.5 \\
\bottomrule
\end{tabular}
\captionof{table}{Ablation of auxiliary losses across the Intent Benchmark, $\mathrm{\tau}^2$-Bench, and BFCL-V3.
Each row removes one loss component while keeping training data, reward, and CM settings identical, and ``$-$all aux'' indicates the simultaneous removal of both auxiliary losses.
Best in \textbf{bold}.}
\label{tab:ablation_loss}
\end{table}

\subsection{Experimental Setup}

\noindent\textbf{Benchmarks.} We evaluate on three benchmarks: (i) the DynamicIntent Benchmark (different intent-fluctuation scenarios, ID and OOD splits with disjoint tool pools, diagnosed by the five metrics in Section~3.2 and three composite scores); (ii) BFCL-V3 (standard function-calling, partial overall accuracy)~\cite{patil2025the}; and (iii) $\mathrm{\tau}^2$-Bench (long-horizon multi-turn, airline and retail domains)~\cite{barres2025tau2benchevaluatingconversationalagents}. All inference uses greedy decoding with thinking disabled (no-think mode).

\noindent\textbf{Baselines.} All methods originate from the same base model, a Qwen3-8B supervised-fine-tuned on the DynamicIntent Dataset. We compare \textbf{Base} (the foundational SFT-only model without reinforcement learning) against seven context-management strategies trained with the PPO pipeline: \textbf{PPO-noCM} retains the full, uncompressed dialogue history as the standard RL baseline; \textbf{FIFO-$k$}~\cite{yang2024sweagentagentcomputerinterfacesenable} serves as a heuristic truncation method that preserves only the most recent $k$ interaction turns; \textbf{Prompt-Comp}~\cite{lee2025learningcontextualizewebpages,wang2026openhandssoftwareagentsdk} acts as a naive generative baseline employing a static context-summarization instruction; \textbf{LLMLingua}~\cite{jiang-etal-2023-llmlingua,pan-etal-2024-llmlingua} applies token-level extractive compression leveraging an encoder-only language model; \textbf{ACON}~\cite{kang2026aconoptimizingcontextcompression} represents a state-of-the-art framework that optimizes compression guidelines in natural language space via failure analysis; \textbf{MEM1}~\cite{zhou2025mem1learningsynergizememory} employs a learnable context compression policy trained jointly with the agent through reinforcement learning; and \textbf{RL-STA}~\cite{chen2026breakingcontextualinertiareinforcement} breaks contextual inertia by incorporating single-turn anchors during reinforcement learning to stabilize multi-turn interactions. Full implementation details are provided in Section~\ref{sec:app_exp}.

\noindent\textbf{Training.} RL is conducted in two stages. Stage~1 trains on single-turn slices to establish tool-selection proficiency with the self-generated CM; Stage~2 seeds from Stage~1 and trains on multi-turn trajectories with the self-generated CM, the hierarchical reward, and the two auxiliary losses. 

\noindent\textbf{Implementation Details.} We use AdamW with a constant actor learning rate of $1{\times}10^{-6}$, a train batch size of 8 with 4 rollouts per prompt, entropy coefficient $0.005$, and a low-variance KL-in-reward term with coefficient $0.001$. Max prompt length is 28{,}672 and max response length is 1{,}024 tokens. The two auxiliary-loss weights are $w_{\text{ext}}{=}0.001$ and $w_{\text{dist}}{=}0.01$. Training runs on 8 GPUs with Ulysses sequence parallelism of size 4, tensor parallelism of size 8, and parameter/optimizer offloading under a dynamic token budget of 30{,}720. Complete hyperparameters and dataset configurations are in Section~\ref{sec:app_exp}.

\subsection{Models and Infrastructure}
\label{sec:app_exp}
We conduct all experiments on an 8B Qwen-based model. Online rollout generation and inference are implemented with the vLLM framework in bfloat16 precision, and PPO-based reinforcement learning is conducted with the \texttt{veRL} framework on $8\times$A800-80GB GPUs. The actor and critic share the base model. The main implementation parameters are summarized in Table~\ref{tab:impl_params}.

\begin{table}[htb]
\centering
\small
\begin{tabular}{lcc}
\toprule
\textbf{Parameter} & \textbf{Stage 1 (single-step)} & \textbf{Stage 2 (multi-turn)} \\
\midrule
Algorithm & \multicolumn{2}{c}{PPO + GAE ($\gamma{=}0.99,\;\lambda{=}0.95$)} \\
Actor learning rate & \multicolumn{2}{c}{$1\times10^{-6}$} \\
Train batch size & \multicolumn{2}{c}{8} \\
PPO mini-batch size & 8 & 2 \\
Rollout responses per prompt ($N$) & \multicolumn{2}{c}{4} \\
Clip ratio (high) & \multicolumn{2}{c}{0.22} \\
KL in reward / coef / type & \multicolumn{2}{c}{enabled / 0.001 / low-variance KL} \\
Entropy coefficient & \multicolumn{2}{c}{0.005} \\
Max prompt / response length & \multicolumn{2}{c}{28{,}672 / 1{,}024} \\
Truncation policy & Left & Error \\
Max user / assistant turns & 1 / 1 & 12 / 12 \\
Tensor parallel (rollout) & \multicolumn{2}{c}{8} \\
Ulysses sequence parallel & \multicolumn{2}{c}{4} \\
Param offload & No & Yes \\
Optimizer offload & \multicolumn{2}{c}{Yes} \\
Gradient checkpointing & \multicolumn{2}{c}{enabled} \\
GPU memory utilization (vLLM) & 0.28 & 0.25 \\
\midrule
Self-generated CM & \cmark & \cmark \\
Auxiliary losses ($w_{\text{ext}}/w_{\text{dist}}$) & --- & \cmark\ ($0.001/0.01$) \\
Reward & $\mathcal{R}_{\text{single}}$  & $\mathcal{R}_{\text{total}}$  \\
\midrule
Layer weights ($\alpha/\beta/\gamma$) & --- & $1.0 / 1.0 / 0.5$ \\
Reward bounds & --- & $[-4.0, +8.0]$ \\
\multicolumn{3}{l}{\textit{Reward hyperparameters (Stage 2 only)}} \\
\multicolumn{3}{l}{\quad JGA: $\frac{n_{\text{hit}}}{n_{\text{calls}}} \times 2.5$ (cap 5.0) \quad SCRR stale: $-0.5$ \quad halluc: $-0.8$} \\
\multicolumn{3}{l}{\quad RIR loop: $-2.0$ \quad RIR invalid: $-1.0$ \quad Progress: $+0.3$/step (cap $0.9$) \quad Prec.: $-0.5$} \\
\multicolumn{3}{l}{\quad DTCR: $+4.0$/$+2.5$/$+1.0$ (full/high/partial) \quad ACR: $\in[-1.0, 1.0]$ \quad ISSR: $+2.0 \times s_{\text{LLM}}$ if $s_{\text{LLM}}>0.3$} \\
\bottomrule
\end{tabular}
\caption{Implementation parameters for the two-stage PPO training. Both stages use the self-generated CM. Stage~1 (single-step) establishes tool-selection proficiency; Stage~2 (multi-turn) adds hierarchical reward and auxiliary losses. Shared settings span both columns.}
\label{tab:impl_params}
\end{table}

\subsection{Two-Stage Training and Data}
IACM-RL is trained in two sequential PPO stages, with Stage~1 initialized from the SFT checkpoint. Stage~1 establishes tool-selection proficiency on 17{,}391 single-turn slices: each sample is a history prefix ending before a gold assistant action, and the model generates exactly one response under the single-scene reward $\mathcal{R}_{\text{single}}$ that scores argument-level correctness. Stage 2 continues leveraging the self-generated Context Manager (CM) with state refresh triggered at every dialogue interaction boundary; it trains over 5{,}639 multi-turn trajectories (each containing up to 12 user utterances and 12 assistant responses), and further optimizes the policy via the hierarchical total reward $\mathcal{R}_{\text{total}}$ alongside the two auxiliary losses. The Stage~1 checkpoint at 600 steps seeds Stage~2, which runs for 100 to 300 steps; checkpoints are saved every 25 steps and the final model is selected by validation performance. This two-stage design isolates ``learning to call tools correctly'' from ``learning to maintain state under dynamic intent fluctuations'', stabilizing credit assignment.

The training data combines proprietary DynamicIntent tool trajectories with two open-source benchmarks (BFCL-V3 and $\mathrm{\tau}^2$-Bench), all converted to a unified RL format via the pipeline in Section~3. A small fraction of pure-linguistic queries without tool calls is included to prevent conversational capability forgetting during RL. Table~\ref{tab:data_comp} reports the per-source composition; the scenario distribution of the Stage~2 multi-turn set is given in Section~\ref{sec:scenario_dist}. For BFCL-V3 and  $\mathrm{\tau}^2$-Bench, all official evaluation instances are excluded. We additionally de-duplicate training and evaluation data at the normalized tool-schema and task-instance levels.

\begin{table}[htb]
\centering
\small
\begin{tabular}{lccc}
\toprule
\textbf{Source} & \textbf{Stage 1 (single-step)} & \textbf{Stage 2 (multi-turn)} & \textbf{Role} \\
\midrule
DynamicIntent & 7{,}899 & 2{,}304 & core intent-fluctuation scenarios \\
BFCL-V3 & 5{,}233 & 1{,}772 & open function-calling diversity \\
$\mathrm{\tau}^2$-Bench & 2{,}609 & 1{,}040 & long-horizon dialogue \\
Language (no tool) & 1{,}650 & 523 & anti-forgetting injection \\
\midrule
Total & 17{,}391 & 5{,}639 & \\
\bottomrule
\end{tabular}
\caption{Training data composition by source. ``DynamicIntent'' denotes proprietary intent-fluctuation trajectories; ``Language'' denotes pure-linguistic queries without tool calls, used as an anti-forgetting injection.}
\label{tab:data_comp}
\end{table}

\begin{table}[htb]
\centering
\begin{minipage}[t]{0.44\linewidth}
\centering
\footnotesize
\begin{tabular}{@{}lccc@{}}
\toprule
\textbf{Dialog Length} & \textbf{PPO-noCM} & \textbf{IACM-RL} & \textbf{$\Delta$} \\
\midrule
Short ($\leq$5k)    & 45.8 & 56.0 & $+$10.2 \\
Medium (5k--11k)    & 35.6 & 36.9 & $+$1.3 \\
Long (11k--16k)     & 8.6 & 23.5 & $+$14.9 \\
Very long ($>$16k)  & 0.0 & 34.8 & $+$34.8 \\
\rowcolor{gray!20}Overall & 28.0 & 35.4 & $\boldsymbol{+}$7.4 \\
\bottomrule
\end{tabular}
\captionof{table}{CM training gain by dialog token length on $\mathrm{\tau}^2$-Bench. $\Delta =$ IACM-RL $-$ PPO-noCM. PPO-noCM struggles on long and complex dialogs, while IACM-RL sustains state tracking via the compact CM block.}
\label{tab:ablation_cm_length}
\end{minipage}\hfill
\begin{minipage}[t]{0.54\linewidth}
\centering
\footnotesize
\begin{tabular}{lccc}
\toprule
\textbf{Scenario Family} & \textbf{PPO-noCM} & \textbf{IACM-RL} & \textbf{$\Delta$} \\
\midrule
\multicolumn{4}{l}{\textit{ID}} \\
Baseline (Simple) & 79.3 & 82.2 & $+$2.9 \\
Modification      & 75.6 & 76.1 & $+$0.5 \\
Clarification     & 67.7 & 69.8 & $+$2.1 \\
Interruption      & 97.0 & 97.6 & $+$0.6 \\
Accumulation      & 66.1 & 67.2 & $+$1.1 \\
Chaining          & 82.4 & 86.7 & $+$4.3 \\
\rowcolor{gray!20}ID Overall & 78.0 & 79.9 & $\boldsymbol{+}$1.9 \\
\midrule
\multicolumn{4}{l}{\textit{OOD}} \\
Modification      & 75.0 & 79.7 & $+$4.7 \\
Clarification     & 66.7 & 70.9 & $+$4.2 \\
Interruption      & 96.3 & 97.0 & $+$0.7 \\
Accumulation      & 65.5 & 64.9 & $-$0.6 \\
Chaining          & 66.1 & 66.1 & $\phantom{+}$0.0 \\
\rowcolor{gray!20}OOD Overall & 73.9 & 75.7 & $\boldsymbol{+}$1.8 \\
\bottomrule
\end{tabular}
\captionof{table}{Per-scenario-family comparison on the Intent Benchmark. Each family uses its primary metric: JGA for Baseline, Modification, and Clarification; ISSR for Interruption; DTCR for Chaining and Accumulation. $\Delta$ is positive when IACM-RL is better.}
\label{tab:ablation_adv_scenario}
\end{minipage}
\end{table}

\subsection{Baselines and Evaluation Protocol}
All methods originate from the same base model, a Qwen3-8B supervised-fine-tuned on the DynamicIntent Dataset. We compare \textbf{Base} (the foundational SFT-only model without reinforcement learning) against seven context-management strategies trained with the PPO pipeline: \textbf{PPO-noCM} retains the full, uncompressed dialogue history as the standard RL baseline; \textbf{FIFO-$k$}~\cite{yang2024sweagentagentcomputerinterfacesenable} preserves only the most recent $k$ turns ($k{=}5$); \textbf{Prompt-Comp}~\cite{lee2025learningcontextualizewebpages,wang2026openhandssoftwareagentsdk} issues a static summarization instruction appended to the system prompt; \textbf{LLMLingua}~\cite{jiang-etal-2023-llmlingua,pan-etal-2024-llmlingua} applies token-level extractive compression with a Qwen3-1.7B model scoring high-perplexity tokens within a budget; \textbf{ACON}~\cite{kang2026aconoptimizingcontextcompression} iteratively refines a natural-language compression instruction via GPT-5.4, triggered when prompt length exceeds a threshold; \textbf{MEM1}~\cite{zhou2025mem1learningsynergizememory} jointly trains a compression policy with the agent via RL, clearing old messages and retaining only the system prompt plus the current turn; and \textbf{RL-STA}~\cite{chen2026breakingcontextualinertiareinforcement} injects single-turn anchor demonstrations during multi-turn RL to stabilize credit assignment and break contextual inertia.

The Intent Benchmark evaluates each of the intent scenarios on approximately 100 dialogs for ID, using a disjoint tool pool for OOD. For \textbf{BFCL-V3}, we evaluate on the single-turn and multi-turn categories. The final score is the unweighted mean of Live, Non-Live, and Multi-turn accuracy (Partial Overall). For $\mathbf{\tau^2}$-\textbf{Bench}, we evaluate airline and retail domains, extracting the Average Reward from the agent performance metrics block. All RL models use greedy decoding with temperature 0.

\subsection{Main Results}

\paragraph{IACM-RL achieves the best overall average with strong tool-calling on complex scenarios.}
Table~\ref{tab:main} shows that IACM-RL attains the highest Avg of 64.0, leading on ID Cognitive, OOD Cognitive, and BFCL, and tying for the best airline score on $\mathrm{\tau}^2$-Bench. Compression baselines such as FIFO-k, Prompt-Comp, and LLMLingua trail on Cognitive because aggressive history truncation discards state needed to track shifting goals. While PPO-noCM benefits from the full reward but lacks explicit state tracking, resulting in uneven gains across metrics.

\paragraph{IACM-RL leads on external benchmarks without sacrificing stability.}
On BFCL-V3, IACM-RL achieves a score of 63.3, demonstrating comparable performance to Base (63.1) and RL-STA (63.2), indicating that the CM does not compromise general tool-calling capabilities. On $\mathrm{\tau}^2$-Bench, IACM-RL ties for the best airline score at 38.0 while compression baselines lag on at least one domain, showing that truncation-based retention is brittle on long-horizon tracking.

\paragraph{IACM-RL demonstrates good out-of-domain generalization.} On the OOD split with a fully disjoint tool pool, IACM-RL maintains a Cognitive score of 36.3, close to its ID score of 36.5. In contrast, Base drops to 33.0 and RL-STA to 33.8 on OOD. By internalizing dynamic state-tracking capabilities during optimization, IACM-RL avoids overfitting to tool-specific surface features. This inherent proficiency allows the agent to robustly manage intent fluctuations and track shifting goals, ensuring stable execution even across entirely unseen tool schemas.

\subsection{Ablation Study and Discussion}
\label{sec:ablation}

\paragraph{Each auxiliary loss contributes to overall stability and performance.}
We train three variants, each removing one auxiliary loss or both simultaneously, while keeping the base model, CM, and hierarchical reward unchanged. Table~\ref{tab:ablation_loss} shows that removing $\mathcal{L}_{\text{ext}}$ degrades OOD Cognitive substantially, from 36.3 to 32.7, confirming that CM Extraction is a crucial driver of self-improving state quality. Removing $\mathcal{L}_{\text{dist}}$ also hurts OOD generalization, dropping to 31.4, as state tracking is no longer distilled into the weights. Removing both reduces the model to pure PPO with self-generated CM, achieving Avg 61.5 versus 64.0 for the full model. It still benefits from the hierarchical reward but loses both the CM-shaping and internalization signals.

\paragraph{IACM-RL demonstrates greater robustness on long dialogs.}
We split $\mathrm{\tau}^2$-Bench trajectories into four buckets by total token length and compare IACM-RL against PPO-noCM in each bucket. Table~\ref{tab:ablation_cm_length} shows that on short dialogs up to 5k tokens, IACM-RL improves reward from 45.8 to 56.0. On long dialogs of 11k--16k, PPO-noCM collapses to 8.6 while IACM-RL reaches 23.5. On very long dialogs above 16k, PPO-noCM scores 0.0 whereas IACM-RL still achieves 34.8, confirming that the compact CM block sustains state tracking where raw-history scanning fails entirely.

\paragraph{CM resistance is strongest in Modification and Clarification scenarios.}
We group the intent scenarios into five families and compare IACM-RL against PPO-noCM using each family's primary metric. Table~\ref{tab:ablation_adv_scenario} shows that on OOD, the gains are largest in Modification with a delta of +4.7 and Clarification with +4.2, consistent with the CM's $\mathcal{C}_{\text{slots}}$ stale flag directly reducing SCRR on modification turns.

\paragraph{IACM-RL retains higher accuracy under adversarial interference.}
We construct adversarial trajectories by inserting chit-chat and fake tasks between the original task turns at four severity levels: L0 clean, L1 with 3 chit-chat messages, L2 with 8, and L3 with 3 fake tasks plus 5 chit-chat messages. Table~\ref{tab:ablation_adv} shows that at L0 both methods are comparable at 81.8\% and 83.2\%. As interference grows, PPO-noCM drops to 25.6\% at L3 while IACM-RL still achieves 36.8\%, an 11.2-point margin, because the CM solidifies $\mathcal{G}_{\text{current}}$ and $\mathcal{C}_{\text{slots}}$ in the system prompt, shielding the policy from mid-context noise.

\paragraph{All three reward layers contribute to performance.}
We train four variants, each disabling one reward layer, namely $R_{\text{belief}}$, $R_{\text{action}}$, or $R_{\text{outcome}}$, or all three for a pure PPO baseline with only format penalty, while keeping the CM and auxiliary losses unchanged. Table~\ref{tab:ablation_reward} shows that removing $R_{\text{belief}}$ causes the largest drop: Avg falls from 64.0 to 57.4 and BFCL from 63.3 to 47.5, because without tool-accuracy supervision the policy overfits to behavioral rewards and loses general calling proficiency. Removing $R_{\text{action}}$ hurts OOD Cognitive, dropping from 36.3 to 31.6, as loop suppression is disabled. Removing $R_{\text{outcome}}$ has a milder effect. Pure PPO still leverages the CM module with Avg 61.0 but lags on Intent metrics.

\begin{table}[htb]
\centering
\footnotesize
\setlength{\tabcolsep}{3.5pt}
\begin{tabular}{lcccccccccc}
\toprule
\multirow{2}{*}{\textbf{Reward Variant}} & \multicolumn{3}{c}{\textbf{Intent ID}} & \multicolumn{3}{c}{\textbf{Intent OOD}} & \multicolumn{2}{c}{\textbf{$\mathrm{\tau}^2$-Bench}} & \multirow{2}{*}{\textbf{BFCL}} & \multirow{2}{*}{\textbf{Avg}} \\
\cmidrule(lr){2-4} \cmidrule(lr){5-7} \cmidrule(lr){8-9}
 & Cog & Beh & Out & Cog & Beh & Out & Air & Ret & & \\
\midrule
\rowcolor{gray!20}\textbf{Full reward} & \textbf{36.5} & \textbf{99.9} & 87.3 & \textbf{36.3} & 99.6 & 81.2 & \textbf{38.0} & \textbf{34.2} & 63.3 & \textbf{64.0} \\
$-R_{\text{belief}}$     & 36.0 & \textbf{99.9} & 69.6 & 33.1 & \textbf{100.0} & 67.1 & 34.0 & 29.8 & 47.5 & 57.4 \\
$-R_{\text{action}}$     & 35.7 & 99.8 & 85.3 & 31.6 & 99.3 & 81.3 & 34.0 & 31.5 & 60.7 & 62.1 \\
$-R_{\text{outcome}}$    & 36.2 & 99.8 & \textbf{87.8} & 33.9 & 99.4 & \textbf{82.1} & 30.0 & 28.9 & 62.7 & 62.3 \\
Format penalty only  & 35.1 & \textbf{99.9} & 83.6 & 33.7 & 99.6 & 78.5 & 26.0 & 28.1 & \textbf{65.3} & 61.0 \\
\bottomrule
\end{tabular}
\caption{
Reward component ablation. Each row disables one reward layer while keeping all other components identical, whereas ``Format penalty only'' denotes the complete removal of all three reward layers. Removing $R_{\text{belief}}$ causes the sharpest BFCL drop from 63.3 to 47.5. Best in \textbf{bold}.
}
\label{tab:ablation_reward}
\end{table}

\paragraph{Accumulate is the best default update policy.}
We separately train three variants with different CM update policies under otherwise identical settings: Accumulate, which incrementally updates the BeliefState each turn as the default; Last-only, which resets the state before each turn; and First-only, which freezes the CM after the first turn. Table~\ref{tab:ablation_temporal} shows that Accumulate achieves the highest Avg at 64.0, while Last-only and First-only trail at 62.4 and 60.1, confirming that retaining prior tool results across turns benefits long-chain and parallel scenarios.

\begin{table}[htb]
\centering
\footnotesize
\setlength{\tabcolsep}{3.5pt}
\begin{tabular}{lcccccccccc}
\toprule
\multirow{2}{*}{\textbf{Update Policy}} & \multicolumn{3}{c}{\textbf{Intent ID}} & \multicolumn{3}{c}{\textbf{Intent OOD}} & \multicolumn{2}{c}{\textbf{$\mathrm{\tau}^2$-Bench}} & \multirow{2}{*}{\textbf{BFCL}} & \multirow{2}{*}{\textbf{Avg}} \\
\cmidrule(lr){2-4} \cmidrule(lr){5-7} \cmidrule(lr){8-9}
 & Cog & Beh & Out & Cog & Beh & Out & Air & Ret & & \\
\midrule
\rowcolor{gray!20}\textbf{Accumulate (OURS)} & 36.5 & \textbf{99.9} & 87.3 & \textbf{36.3} & 99.6 & 81.2 & \textbf{38.0} & \textbf{34.2} & \textbf{63.3} & \textbf{64.0} \\
Last-only            & 36.4 & \textbf{99.9} & 87.2 & 33.8 & \textbf{99.8} & \textbf{83.0} & 30.0 & 29.8 & 61.7 & 62.4 \\
First-only           & \textbf{38.7} & 99.8 & \textbf{87.4} & 35.1 & 99.4 & 82.6 & 28.0 & 8.8 & 61.3 & 60.1 \\
\bottomrule
\end{tabular}
\caption{CM temporal alignment ablation. Each variant is trained separately with a different CM update policy under otherwise identical settings.}
\label{tab:ablation_temporal}
\end{table}

\paragraph{LLM simulation is a faithful proxy for real tool execution.}
We train two SFT models on Qwen3-8B with identical settings except the tool-return source: one uses LLM-simulated returns, the other uses real MCP execution. Table~\ref{tab:mcp_vs_sim} shows that on Intent, the two models are nearly identical in Cog and Beh. On $\mathrm{\tau}^2$-Bench, they are comparable, with real-MCP slightly better on airline and LLM-sim better on retail. These results validate that LLM-driven simulation is a scalable proxy that matches real execution across all core metrics.

\begin{table}[htb]
\centering
\footnotesize
\setlength{\tabcolsep}{3.5pt}
\begin{tabular}{lccccccccc}
\toprule
\multirow{2}{*}{\textbf{Tool Source}} & \multicolumn{3}{c}{\textbf{Intent ID}} & \multicolumn{3}{c}{\textbf{Intent OOD}} & \multicolumn{2}{c}{\textbf{$\mathrm{\tau}^2$-Bench}} & \multirow{2}{*}{\textbf{Avg}} \\
\cmidrule(lr){2-4} \cmidrule(lr){5-7} \cmidrule(lr){8-9}
 & Cog & Beh & Out & Cog & Beh & Out & Air & Ret & \\
\midrule
LLM-sim  & 39.5 & \textbf{99.8} & 87.0 & 35.8 & 99.2 & \textbf{85.2} & 17.0 & \textbf{18.4} & \textbf{60.2} \\
Real MCP & \textbf{39.6} & \textbf{99.8} & \textbf{87.4} & \textbf{35.9} & \textbf{99.8} & 83.6 & \textbf{20.0} & 14.9 & 60.1 \\
\bottomrule
\end{tabular}
\caption{LLM simulation vs. real MCP execution. Both models are SFT-only on Qwen3-8B with identical settings except tool-return source.}
\label{tab:mcp_vs_sim}
\end{table}

\paragraph{IACM-RL is stable across random seeds.}
We train IACM-RL with three random seeds and report the full benchmark results in Table~\ref{tab:ablation_seed}. The variance is small across all 9 metrics, confirming stable reproducibility.

\begin{table}[htb]
\centering
\footnotesize
\setlength{\tabcolsep}{3pt}
\begin{tabular}{lcccccccccc}
\toprule
\multirow{2}{*}{\textbf{Method}} & \multicolumn{3}{c}{\textbf{Intent ID}} & \multicolumn{3}{c}{\textbf{Intent OOD}} & \multirow{2}{*}{\textbf{BFCL}$\uparrow$} & \multicolumn{2}{c}{\textbf{Tau2}$\uparrow$} & \multirow{2}{*}{\textbf{Avg}} \\
\cmidrule(lr){2-4} \cmidrule(lr){5-7} \cmidrule(lr){9-10}
 & Cog & Beh & Out & Cog & Beh & Out & & Air & Ret & \\
\midrule
seed=42 & 36.5 & 99.9 & 87.3 & 36.3 & 99.6 & 81.2 & 63.3 & 38.0 & 34.2 & 64.0 \\
seed=123 & 36.8 & 99.9 & 84.5 & 34.3 & 100.0 & 78.9 & 62.6 & 38.0 & 36.8 & 63.5 \\
seed=7 & 36.5 & 99.9 & 86.4 & 32.4 & 98.5 & 81.7 & 63.6 & 40.0 & 30.7 & 63.3 \\
\rowcolor{gray!20}\textbf{mean$\pm$std} & 36.6$\pm$0.2 & 99.9$\pm$0.0 & 86.1$\pm$1.4 & 34.3$\pm$2.0 & 99.4$\pm$0.8 & 80.6$\pm$1.5 & 63.2$\pm$0.5 & 38.7$\pm$1.2 & 33.9$\pm$3.1 & 63.6$\pm$0.4 \\
\bottomrule
\end{tabular}
\caption{Random seed variance across all benchmarks (mean$\pm$std over 3 seeds). Avg is the unweighted mean over the 9 columns.}
\label{tab:ablation_seed}
\end{table}

\paragraph{The default loss weights are well-tuned.}
We scale both auxiliary-loss weights simultaneously by 0.25, 0.5 and 2.0 relative to the default, while keeping the CM and hierarchical reward unchanged. Table~\ref{tab:ablation_weight} shows that the default weights achieve the best Avg at 64.0. Halving the weights to 0.5$\times$ reduces Avg to 62.7, while quartering to 0.25$\times$ drops further to 62.1, as the CM-shaping and internalization signals become too weak to guide the policy effectively. Doubling to 2$\times$ also degrades performance to 62.6, as overly strong auxiliary gradients destabilize the PPO update and interfere with the policy-gradient signal. These results confirm that the default weight setting strikes the right balance between the auxiliary losses and the main PPO objective.

\begin{table}[htb]
\centering
\footnotesize
\setlength{\tabcolsep}{3.5pt}
\begin{tabular}{lcccccccccc}
\toprule
\multirow{2}{*}{\textbf{Scale}} & \multicolumn{3}{c}{\textbf{Intent ID}} & \multicolumn{3}{c}{\textbf{Intent OOD}} & \multicolumn{2}{c}{\textbf{$\mathrm{\tau}^2$-Bench}} & \multirow{2}{*}{\textbf{BFCL}} & \multirow{2}{*}{\textbf{Avg}} \\
\cmidrule(lr){2-4} \cmidrule(lr){5-7} \cmidrule(lr){8-9}
 & Cog & Beh & Out & Cog & Beh & Out & Air & Ret & & \\
\midrule
\rowcolor{gray!20}\textbf{1.0$\times$ (default)} & 36.5 & \textbf{99.9} & \textbf{87.3} & \textbf{36.3} & \textbf{99.6} & 81.2 & \textbf{38.0} & \textbf{34.2} & \textbf{63.3} & \textbf{64.0} \\
0.25$\times$ & 35.9 & 99.8 & 86.5 & 34.1 & 99.4 & \textbf{84.4} & 26.0 & 30.7 & 62.4 & 62.1 \\
0.5$\times$ & \textbf{36.8} & 99.8 & 86.4 & 33.7 & 98.9 & 83.0 & 34.0 & 29.8 & 61.8 & 62.7 \\
2.0$\times$ & 35.9 & 99.8 & 87.2 & 34.2 & 99.5 & 84.2 & 28.0 & 32.5 & 62.4 & 62.6 \\
\bottomrule
\end{tabular}
\caption{Auxiliary loss weight sensitivity. Both weights $w_{\text{ext}}$ and $w_{\text{dist}}$ are scaled simultaneously. The default setting achieves the best Avg; both reducing and amplifying the weights degrade performance.}
\label{tab:ablation_weight}
\end{table}

To test whether IACM-RL transfers to a smaller backbone, we repeat the main experiment on a 1.7B Qwen model with the same two-stage training and all components (CM, hierarchical reward, auxiliary losses) kept identical. Table~\ref{tab:main_1_7b} reports the results. As expected, the 1.7B model exhibits substantially lower capacity, leading to absolute score drops across all methods. Despite this, IACM-RL still achieves the highest overall average score (54.2) and leads in key metrics including ID Behavioral, ID Outcome, BFCL, and both domains of $\mathrm{\tau}^2$-Bench. However, it no longer dominates every cognitive and outcome sub-metric (e.g., trailing MEM1 in OOD Cognitive and PPO-noCM in OOD Outcome), suggesting that the full potential of explicit Context Management relies on a baseline level of model reasoning capacity. Notably, learnable compression via MEM1 demonstrates strong robustness at this scale, ranking second overall (53.0), whereas naive truncation methods like FIFO-k suffer severe degradation (47.0).

\begin{table}[htb]
\centering
\footnotesize
\setlength{\tabcolsep}{3.5pt}
\begin{tabular}{lcccccccccc}
\toprule
\multirow{2}{*}{\textbf{Method}} & \multicolumn{3}{c}{\textbf{Intent ID}} & \multicolumn{3}{c}{\textbf{Intent OOD}} & \multicolumn{2}{c}{\textbf{$\mathrm{\tau}^2$-Bench}} & \multirow{2}{*}{\textbf{BFCL}} & \multirow{2}{*}{\textbf{Avg}} \\
\cmidrule(lr){2-4} \cmidrule(lr){5-7} \cmidrule(lr){8-9}
 & Cog & Beh & Out & Cog & Beh & Out & Air & Ret & & \\
\midrule
Base         & 16.0 & 99.3 & 56.3 & 25.1 & \textbf{99.6} & 65.8 & 22.4 & 5.3 & 48.5 & 48.7 \\
PPO-noCM     & 17.8 & 99.6 & 70.0 & 23.0 & 99.1 & \textbf{75.0} & 26.0 & 5.3 & 49.1 & 51.6 \\
FIFO-k       & 7.3 & 92.5 & 65.6 & 15.2 & 93.2 & 68.7 & 24.0 & 6.1 & 50.1 & 47.0 \\
Prompt-Comp  & 12.9 & 99.7 & 62.5 & 21.4 & 98.8 & 72.3 & 22.0 & 9.7 & 47.5 & 49.6 \\
ACON         & 20.1 & 99.3 & 61.8 & 23.8 & 99.5 & 67.1 & 32.0 & 4.4 & 49.7 & 50.9 \\
LLMLingua    & \textbf{21.6} & 99.9 & 70.7 & 25.3 & 99.3 & 74.4 & 14.0 & 6.1 & 47.8 & 51.0 \\
MEM1         & 19.7 & 99.7 & 67.9 & \textbf{27.9} & \textbf{99.6} & 71.1 & \textbf{34.0} & 7.0 & 50.1 & 53.0 \\
RL-STA       & 17.9 & 99.9 & 60.3 & 24.5 & 98.7 & 65.2 & 28.0 & 4.4 & 50.2 & 49.9 \\
\rowcolor{gray!20}\textbf{IACM-RL (Ours)} & 17.3 & \textbf{100.0} & \textbf{71.4} & 25.8 & 99.4 & 74.3 & \textbf{34.0} & \textbf{14.9} & \textbf{50.4} & \textbf{54.2} \\
\bottomrule
\end{tabular}
\caption{Main results on a 1.7B Qwen model. All methods use the same 1.7B backbone, training data, and evaluation method as their corresponding 8B configurations, while retaining their method-specific context-management components. IACM-RL achieves the highest Avg at 54.2, leading in ID Behavioral, BFCL, and $\mathrm{\tau}^2$-Bench. Learnable compression (MEM1) shows strong robustness at this scale, ranking second overall with an Avg of 53.0.}
\label{tab:main_1_7b}
\end{table}

\begin{table}[htb]
\centering
\begin{minipage}[t]{0.48\linewidth}
\centering
\footnotesize
\begin{tabular*}{\columnwidth}{@{\extracolsep{\fill}}lccc@{}}
\toprule
\textbf{Interference level} & \textbf{PPO-noCM} & \textbf{IACM-RL} & \textbf{$\Delta$} \\
\midrule
L0\ clean                   & 81.8 & 83.2 & $+$1.4  \\
L1\ 3 chit-chat             & 34.1 & 50.7 & $+$16.6 \\
L2\ 8 chit-chat             & 27.8 & 40.4 & $+$12.6 \\
L3\ 3 fake $+$ 5 chit-chat  & 25.6 & 36.8 & $+$11.2 \\
\bottomrule
\end{tabular*}
\captionof{table}{Adversarial drift robustness (tool-call accuracy, \%) on 500 sampled trajectories with inserted interference. L0 is clean; L1/L2 insert 3/8 chit-chat messages; L3 inserts 3 fake tasks plus 5 chit-chat messages between the original task turns. $\Delta$ is positive when IACM-RL outperforms PPO-noCM.}
\label{tab:ablation_adv}
\end{minipage}\hfill
\begin{minipage}[t]{0.48\linewidth}
\centering
\footnotesize
\begin{tabular}{lccc}
\toprule
\textbf{Char length} & \textbf{wo CM} & \textbf{w/ CM} & \textbf{$\Delta$} \\
\midrule
Short ($<$5K)       & 0.615 & 0.667 & $+$0.052 \\
Medium (5K--15K)    & 0.216 & 0.500 & $+$0.284 \\
Long (15K--30K)     & 0.086 & 0.397 & $+$0.311 \\
Very long ($>$30K)  & 0.000 & 0.202 & $+$0.202 \\
\rowcolor{gray!20}Overall & 0.245 & 0.323 & $\boldsymbol{+}$0.078 \\
\bottomrule
\end{tabular}
\captionof{table}{Inference-time CM analysis on $\mathrm{\tau}^2$-Bench airline (Qwen3-8B, no RL training), bucketed by conversation character length. $\Delta =$ w/ CM $-$ wo CM. The CM gain grows with dialogue length; above 30K, wo CM collapses to 0 while CM still achieves 0.202.}
\label{tab:ablation_cm_inference}
\end{minipage}
\end{table}

\subsection{Inference-Time CM Analysis on $\mathrm{\tau}^2$-Bench}

\paragraph{The Context Manager provides increasing inference-time gains with longer dialogues.}
To isolate the inference-time benefit of the self-generated CM block from training, we evaluate the same Qwen3-8B base model (no RL training) under two pipelines on $\mathrm{\tau}^2$-Bench airline: \textbf{wo CM}, where the model receives the raw dialogue history, and \textbf{w/ CM}, where a CM proxy injects the Context Manager block into the system prompt before each assistant turn.
Table~\ref{tab:ablation_cm_inference} groups trajectories by conversation character length. On short dialogues below 5K, CM yields a marginal gain of $+0.052$. As length increases, the gain grows to $+0.284$ on medium (5K--15K) and $+0.311$ on long (15K--30K) dialogues. Above 30K, the no-CM pipeline collapses entirely, while CM still achieves 0.202. This trend confirms that the CM block provides a structural anchor that relates to dialogue complexity: short interactions barely benefit, but under long-horizon intent fluctuations the explicit state summary becomes more valuable.

\section{Attention Visualization}
\label{sec:app_attn}

Figure~\ref{fig:attn_full} visualizes the token-level attention distribution on representative Modification turns for two models: PPO-noCM, which scans the raw dialogue history, and IACM-RL, which conditions on the self-generated CM block shown in Figure~\ref{fig:cm_full}.

For PPO-noCM, attention is uniformly diluted across the long raw history. The tokens carrying the current goal and the latest parameter change receive only a small fraction of the total attention budget, as obsolete constraints and injected chit-chat compete for the same attention mass. This dilution subverts the data dependencies among multiple tools, trapping the agent in catastrophic intent deviation and infinite API loops.

In contrast, IACM-RL concentrates its attention on the CM block in the system prompt, which distills the turn-critical state into a compact representation. The key tokens corresponding to the current goal, confirmed slots, and the override instruction excluding the obsolete jacket exchange receive sharply higher attention weights, while mid-context noise from chit-chat and prior modifications is effectively suppressed. This focused attention pattern confirms that the CM acts as an attention anchor, directing the policy toward the state-relevant information rather than re-deriving it from the verbose history.

\begin{figure}[htb]
\centering
\includegraphics[width=0.82\textwidth]{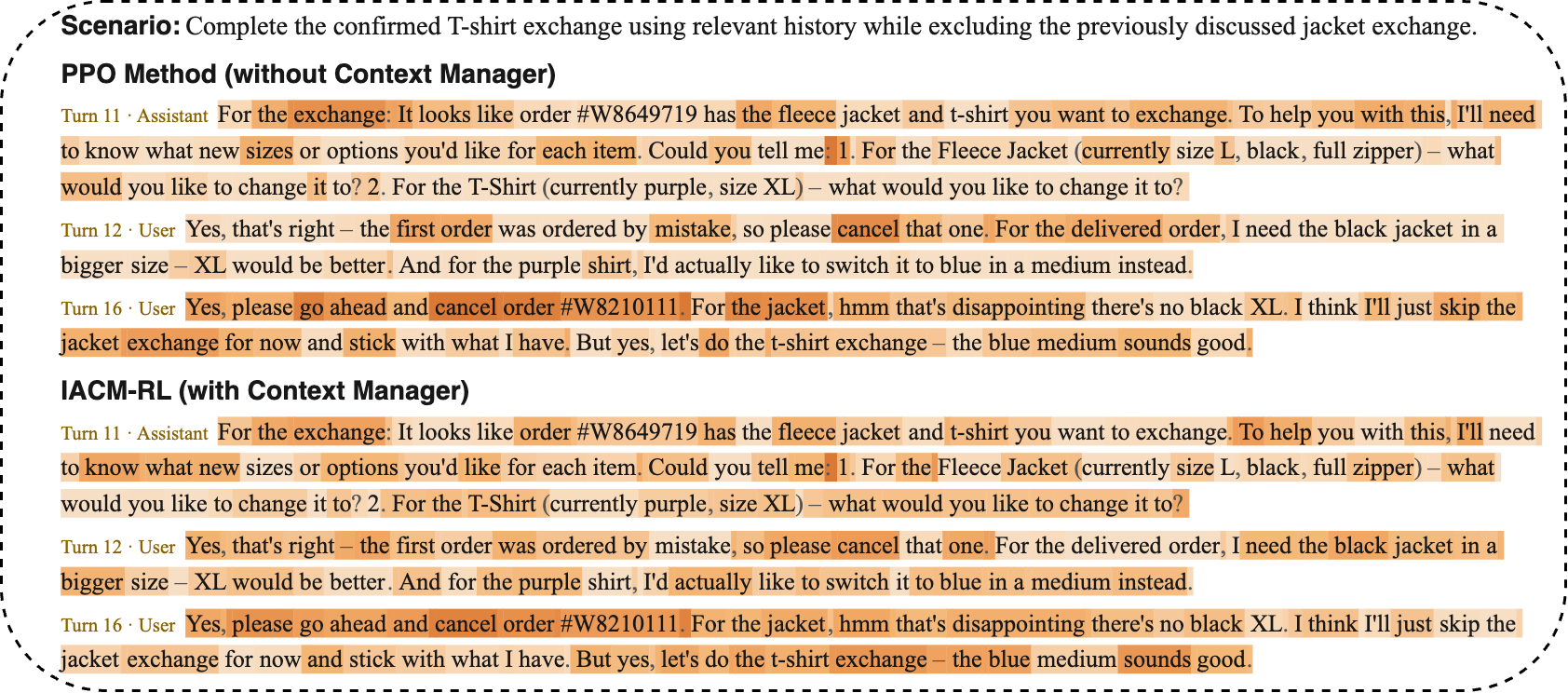}
\caption{Token-level attention comparison on Modification turns. Top: PPO-noCM distributes attention uniformly across the raw history, failing to focus on the current goal and updated parameters. Bottom: IACM-RL uses CM guidance to identify the current execution boundary, confirmed slots, and the override instruction, thereby enhancing attention to these elements while suppressing mid-context noise. Darker colors indicate higher attention.}
\label{fig:attn_full}
\end{figure}

\begin{figure}[htb]
\centering
\includegraphics[width=0.82\textwidth]{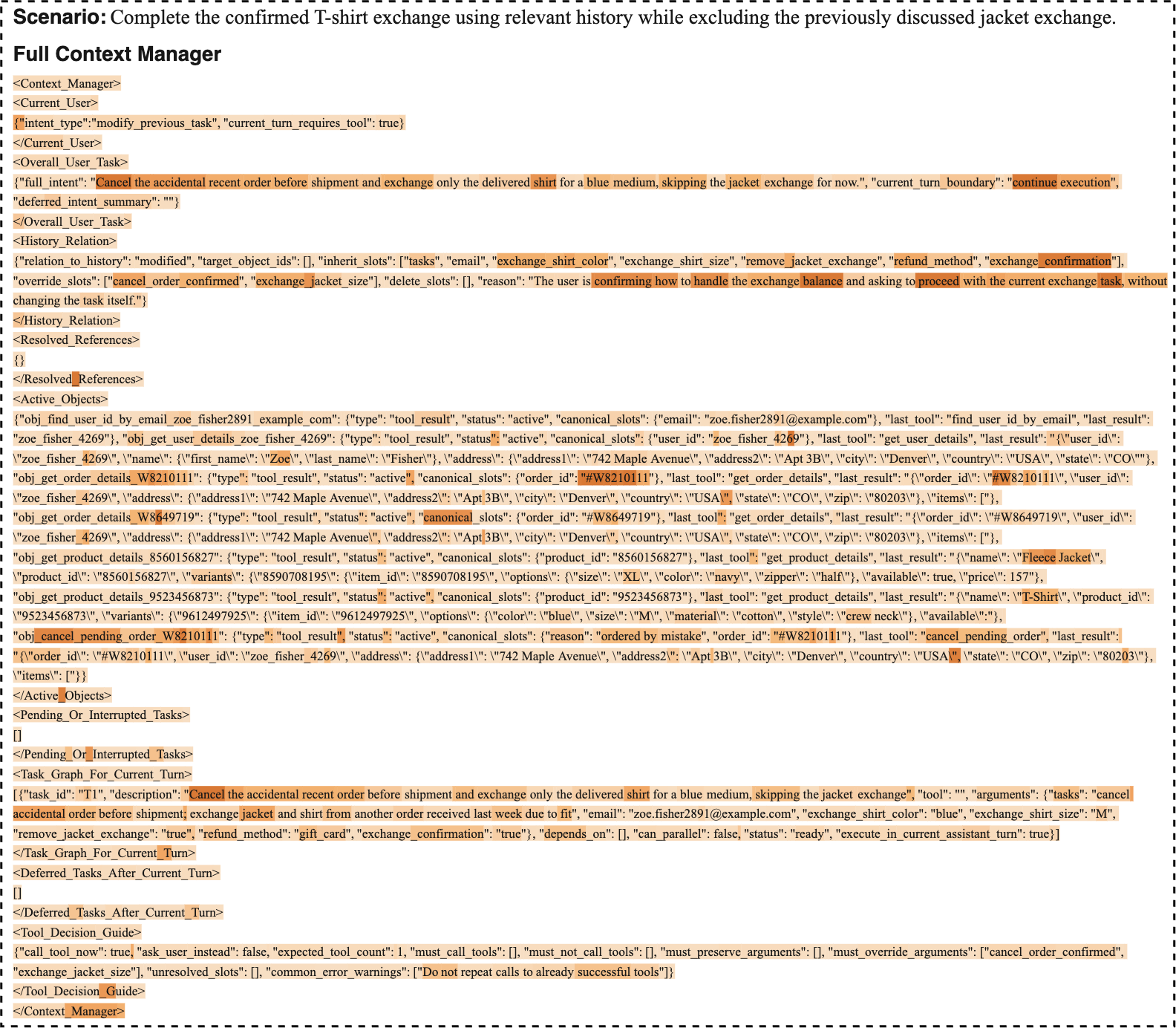}
\caption{Token-level attention over the full Context Manager. IACM-RL selectively focuses on action-relevant CM fields, including the updated current goal, confirmed exchange parameters, the flag that removes the obsolete jacket exchange, and execution guidance, which facilitates the capture of key information from the interaction history. Darker colors indicate higher attention.}
\label{fig:cm_full}
\end{figure}



\section{Conclusion}

We introduce IACM-RL, a framework that explicitly decouples state tracking from action generation for robust tool invocation under dynamic intent fluctuations. At its core is an autoregressive Self-Generated Context Manager governed by a BeliefState, which proactively tracks shifting goals and marks overwritten parameters with a structural stale flag to prevent regression to obsolete values. To internalize this capability, we couple the CM with a hierarchical intent-driven reward and two auxiliary losses, optimized over the DynamicIntent Dataset spanning several fluctuation scenarios. Experiments on the DynamicIntent Benchmark, BFCL-V3, and $\mathrm{\tau}^2$-Bench show that IACM-RL achieves the highest overall average, demonstrates strong out-of-domain generalization, and ablations confirm each auxiliary loss and the CM contribute non-overlapping gains, offering a practical route toward reliable long-horizon agents under noisy, evolving intent.

\section*{Limitations}

All experiments use text-based API tools simulated by LLM; whether the observed gains transfer to multimodal or non-tool agentic settings remains an open question.

\section*{Ethical Considerations}

IACM-RL targets tool-calling agents that must stay reliable under shifting user intent; because agents of this kind could in principle be repurposed to act on a user's behalf without supervision, we confine the present study to a simulated execution environment and recommend that deployments gate irreversible actions behind explicit user confirmation. Test-set annotation was performed by an in-house team at the authors' institution, with no external participants and no personal data collected. All training and evaluation use $8\times$A800-80GB GPUs under a short two-stage schedule.

\section*{Artifact Documentation, Licensing, and Data Statements}
\label{sec:artifact}

\noindent\textbf{Domain and language.} The DynamicIntent Dataset covers personal-assistant tool use---calendar and reminders, messaging and calls, navigation and travel, finance and payments, documents and OCR, media and device control. Dialogue turns, tool schemas, and the Context Manager sub-block tags are in English and Chinese.

\noindent\textbf{Licensing.} The DynamicIntent Dataset and all models we train are our own artifacts. The two open-source benchmarks used for training and evaluation, BFCL-V3~\cite{patil2025the} and $\mathrm{\tau}^2$-Bench~\cite{barres2025tau2benchevaluatingconversationalagents}, are used under their original licenses.

\noindent\textbf{Personally identifiable information.} Every trajectory is synthesized by LLMs and contains no real user records; the profile fields used by the Personalized Information Update scenario are fabricated, and no personal data was collected from annotators.

\noindent\textbf{Offensive content.} The corpus consists of task-oriented assistant dialogues and is not constructed to contain offensive, hateful, or otherwise harmful content.

\bibliography{aaai2027}

\endgroup

\renewcommand{\topfraction}{0.85}
\renewcommand{\bottomfraction}{0.70}
\renewcommand{\textfraction}{0.30}
\renewcommand{\floatpagefraction}{0.85}
\setcounter{topnumber}{4}
\setcounter{bottomnumber}{4}
\setcounter{totalnumber}{6}

\clearpage
\bibliography{aaai2027}
\clearpage

\begingroup
  \let\maketitle\relax
  \RenewEnviron{abstract}{}
  \let\bibliography\SkipImportedBibliography

\maketitle

\begin{abstract}
Executing long-horizon tool invocations in real-world environments is severely challenged by dynamic user intent noise. Existing methods attempt robustness via implicit history scanning or text compression, yet predominantly assume perfect instructions in simplistic scenarios. Inevitably, under fluctuating contexts, obsolete constraints dilute model attention, triggering intent deviation and repetitive API loops. To resolve this, we propose \textbf{IACM-RL}, a comprehensive framework for robust tool invocation. First, we introduce the \textbf{DynamicIntent} pipeline, synthesizing trajectories across several fine-grained fluctuation scenarios, paired with a five-dimensional diagnostic metric suite. Second, IACM-RL deploys a BeliefState-based Self-Generated Context Manager that proactively tracks shifting goals and isolates overwritten parameters using structural stale flags. To autonomously internalize this state-tracking capability, we optimize the policy using a hierarchical intent-driven reward alongside two auxiliary losses (CM extraction and state distillation). Experiments on DynamicIntent, BFCL-V3, and $\mathrm{\tau}^2$-Bench demonstrate that IACM-RL outperforms  baselines, reducing infinite loops and stale context errors while enhancing out-of-domain generalization.
\end{abstract}

\appendix

\section{Prompt Templates}
\label{sec:app_prompts}

This appendix lists the principal prompts used across the DynamicIntent data construction pipeline, the self-generated Context Manager, and the reward/evaluation LLM-as-a-judge modules. Chinese prompts are presented verbatim; English judge prompts follow their original wording. Placeholders in braces are filled at runtime.

\subsection{Data Construction Prompts}
\label{app:prompt_data}

The DynamicIntent pipeline is execution-free: natural-language user queries, assistant tool calls, and mock tool returns are all synthesized by LLMs over the Tool Dependency Graph. Some of the templates below are shown in abbreviated form; the full templates are given in the released code. The core primitive is a back-and-forth translation over each function in a sampled Function Sequence Pattern (FSP): a user query is generated from the schema (back-translation), an assistant tool call is generated from the query (forth-translation), a mock tool return is produced by an LLM simulator, and the assistant then summarizes the return. Intent-fluctuation methods further rewrite the query to inject modifications, interruptions, clarifications, and accumulations.

\noindent\textbf{Tool dependency graph construction.} The Tool Dependency Graph $G=(V,E)$ is built by a multi-source LLM voting: for each ordered tool pair, several LLM accounts independently judge whether the output of the source tool can serve as (part of) the input of the target tool; an edge is added only when a majority agrees across two voting rounds (static schema judgment and input/output-example judgment), and the dependent parameter names are intersected across votes. This graph is the structural basis for long-chain (Chaining) scenarios.
\begin{promptbox}
You will be given a source API function and a target API function. You will also be given an input example and output example of source API function. Your task is to:
1. Judge if the target api is related to the source API.
2. Judge which input parameters of the target API is dependent on the output of source api.

We say one function is related to the source API if:
1) the output of the source API is the premise of executing the function. For example, the output of fileexists('file.txt') API determines whether we can call downloadfile('file.txt').
2) the output of the source API is exactly the input parameters of the function. For example, when calculating the area of a circle, the function getradius(obj) is the source node and calculate(radius) is the target node.
3) the output of the source API is partial input parameters of the function. For example, when posting something to social media, one might first get the content. In this case, the content = getcontent('file.txt') is the source node and posting(content, id, tags) is the target node.

We say one parameter is dependent on the source API if:
1) the output of the source API is exactly the input parameters of the function.
2) the output of the source API is partial input parameters of the function.

Notice that the relation might cross the boundary of domains. For example, when the given APIs are in the domain of weather and travel, it is possible that a weather API could be dependent on a travel API since the weather determines the travel schedule.

Attention:
1. You need to first analyze what the source API function's output is.
2. When evaluating the output of the source API function, first refer to the output examples; the description of source API function are only secondary references.
3. In case of any conflict or invalid output examples, rely on the output examples.
4. The target API function is related to the source API function, but this does not mean that the target API function necessarily has parameters that is dependent on the source API function.
5. In <think> and </think>, give a brief explanation on how you think and make judge. In <judge> and </judge>, if the target api is related to the source API, output yes, otherwise output no, Use lower case. In <parameters> and </parameters>, give a list of parameter names of the target API which are dependent on the source API if any, like ['param1', 'param2', ...], otherwise output a blank list.

Source API: {source_api}
Target API: {target_api}
Input example: {input_example}
Output example of the input example: {output_example}
\end{promptbox}

\noindent\textbf{LLMSimBackend (mock tool execution).} Because the pipeline is execution-free, a dedicated LLM acts as an API server: given a tool's schema, examples, and a concrete call input, it crafts a JSON response that aligns with the API's intended output. This produces the mock tool returns that populate the expected call records and the conversation history. The response is constrained to 100--200 words with rich, practical content.
\begin{promptbox}
Imagine you are an API Server operating within a specialized tool, which contains a collection of distinct APIs. Your role is to deeply understand the function of each API based on their descriptions in the API documentation. As you receive specific inputs for individual API calls within this tool, analyze these inputs to determine their intended purpose. Your task is to craft a JSON formatted response that aligns with the expected output of the API, guided by the provided examples.

Your responses must adhere to a specific JSON structure:
{
    "error": "",
    "response": "<Your_Response>"
}

The error field should remain empty. The response field should contain the content you formulate based on the API's functionality and the input provided. Ensure that your responses are meaningful, directly addressing the API's intended functionality.

Note that:
- your response should be around 100 to 200 words, containing rich information given the api input parameters. Keep your answer short and simple.
- your response must be effective and have practical content.
- if the api response example is null or ineffective, ignore the example and give your independent response.
- you will also receive a conversation history, which includes previous exchanges between the user and the assistant for your reference.

API Documentation: {api_doc}
API Examples: {api_example}
API Input: {tool_input}
Conversation History: {history}
\end{promptbox}

\noindent\textbf{Back-translation (user query generation).} Given the conversation history and the candidate function schema, the LLM role-plays as the user and generates a natural query that invokes the target function. History outputs are referred to by deictic references (e.g., ``the location you just found'') rather than literal values, encouraging multi-turn dependency.
\begin{promptbox}
Now you are role-playing as a user that involves in a multi-turn conversation with a function-calling agent. You will be given the history of this multi-turn conversation, indicated by round numbers. You will also be provided with a list of candidate functions that can be called in this round. I would like you to generate the query of this round which calls one or multiple functions from the candidate function list.

Rules:

- The preferred this round query should be motivated by the outputs from the last round. Preferably, those outputs are used as the input parameters for as least one of the functions being called at this round.
- You should NOT mention which functions to use in your query explicitly.
- Use no parameters besides the parameters indicated in the function documentation.
- After you decide on which function to use, make sure your new query contains information for parameters of the functions you want to call.
- For the parameters from the last round, try not to mention the exact parameters that you will use. Instead, use references such as 'the location you just found', 'With the listed items'... to refer to the output of last round that will be leveraged next.
- For the parameters in the default parameter list, you can use them directly or you can generate new parameter values yourself.
- Try to make the conversation as natural as possible. Mind the logic between two consecutive queries. Do not just create an independent new query.
- Generate the query between <answer> and </answer>.
- If the query contains any relative time expression (e.g., today, tomorrow, the day after tomorrow, next Friday, this weekend), you MUST include an explicit absolute date for that time.
- Allowed date expression styles: either relative+absolute form (e.g., tomorrow (2025.08.20)) or absolute-only form (e.g., 2025.08.20).
- Do NOT output unresolved placeholders or template markers for dates (e.g., <specific date>, <the concrete date of next Friday>).
[History]:{history}
[Candidate Functions]:{candidate_functions}
[Default Parameters]{default_param}

[Output]:
\end{promptbox}

\noindent\textbf{Forth-translation (assistant tool call generation).} The LLM role-plays as the function-calling agent, deriving arguments from the user query and referencing prior assistant outputs. The response follows a fixed tool-call and answer format with a JSON array of function calls.
\begin{promptbox}
Now you are role-playing as a function-calling agent that involves in a multi-turn conversation with a user. You will be given the history of this multi-turn conversation, indicated by round numbers. You will also be provided with a list of candidate functions that can be called in this round. I would like you to generate the function call for this round using this function signature. Make sure the parameters for this candidate function should be derived from the user query and reference outputs from the history.

Rules:
- You should use the function with the original name without any changes.
- For all the functions, make sure your generated function calls contain ALL the required parameters fields from the function documentation. You may also include some optional parameters. However, do not hallucinate any parameters outside of those. Use only the parameters indicated in the required and optional fields of the function documentation.
- Then, the parameter values for the new function should be derived from the user query and must reference the outputs of the assistant.
- You can have parallel function call with the candidate function, i.e., call the function with different set of parameters, for your new query. However, do not call more than three parallel functions.

Format:
Thought: ... <think> ... </think>
Answer: a JSON array of {"name","arguments"} objects inside <answer> ... </answer>.

[History]: {history}
[Candidate Functions]: {candidate_functions}
[User Query]: {query}
\end{promptbox}

\noindent\textbf{Tool-result summarization.} After the LLM simulator produces a mock tool return, the assistant summarizes it into a natural, user-facing response without exposing technical details (function names, JSON, error codes).
\begin{promptbox}
Now you are role-playing as a user-facing AI Assistant. Your primary job is to interpret the results of a tool call that was just executed in the background and formulate a natural, helpful, and user-friendly response. You are the final bridge between the system's actions and the user's understanding.

Rules:
- Your response must be based on the provided tool output. Do not hallucinate information that is not present in the [Last Round Tool Output].
- You must not expose any technical details to the user. Never mention function names, JSON structures, or technical error codes. Your role is to translate these technical details into plain language.
- Your response should directly address the user's last query, which can be inferred from the [History].
- Handle different outcomes gracefully: present data on success, suggest alternatives on empty results, apologize on failure.
- Your tone should be helpful and conversational.
\end{promptbox}

\noindent\textbf{Modification method (parameter rewrite).} To synthesize the ``Modification'' fluctuation, two conflicting user sentences are merged into one that negates the first and adopts the second, simulating a user changing their mind (e.g., ``I want Beijing, oh wait, Shanghai instead'').
\begin{promptbox}
You are an expert AI assistant specializing in Natural Language Understanding and task consolidation. Your primary function is to synthesize list of two sentences into a single, coherent, and natural-sounding final sentence. You must strictly adhere to the rules and process the context provided below.

You will be given four pieces of information:
- Conversation History: The turn-by-turn interaction between a user and an assistant.
- Sentence List: A list of sentences which need to be synthesized. Include sentence one and sentence two.

Your task is to merge all user inputs from the Sentence List into a single sentence. You must follow these rules without exception:
- Preserve Critical Information: The final sentence must contain all critical information of each sentence of the list.
- Appropriate Omission: For repetitive or redundant information, use omission. For information that has appeared in Conversation History, also use omission or referential expressions. Omit everything as long as you think it redundant.
- Ensure Natural Fluency: The merged sentence must be grammatically correct, coherent, and sound like a single, natural request from a human. Avoid robotic phrasing or awkward concatenations.
- Provide Transition: Simulate the user's change of mind and regretful statements. For the two sentences in the list, if the information in the second sentence conflicts with that in the first sentence, you need to negate the conflicting content of the first sentence and adopt the content of the second sentence. For example: I want to go to the one in Beijing---oh, no, it should be the one in Shanghai.
- Produce only the final merged sentence as a single string of text between <answer> and </answer>. Do not include any explanations, introductory phrases, or markdown formatting.

Here are the information:

[Conversation History]:{history}
[Sentence List]:{sentence_list}
\end{promptbox}

\noindent\textbf{Interruption method (function switch).} To synthesize ``Interruption'', the current query is rewritten to make the transition from the previous query more natural, while retaining the original meaning and all information.
\begin{promptbox}
You are a professional user query rewriting expert. Next, you will be given the current query and the previous query. Please rewrite the current query to make the transition with the previous query more natural.
- You must retain the original meaning and all information of the current query.
- Output your answer between <query> and </query>

Previous query:
{last_round_query}
Current query:
{current_query}
\end{promptbox}

\noindent\textbf{DynamicMockTool fallback (training-time).} During RL rollout, when the model's tool call does not match the ground-truth expected calls, a fallback LLM simulates a realistic tool response based on the schema and the actual arguments, so that the trajectory can continue. It returns errors only for two cases, tool does not exist or undefined parameter, and otherwise produces a normal realistic result; this keeps the multi-turn loop from collapsing on out-of-distribution calls.
\begin{promptbox}
You are a tool server that responds to various tool call requests. Your role is to:
- Deeply understand the tool and its format based on its schema
- Process incoming tool calls with their parameters
- Generate appropriate responses based on the tool's expected output
- Only return errors for two specific cases: tool does not exist, or undefined parameter is provided
- For all other cases (including missing required parameters, type mismatches, format errors, invalid enum values), return a normal, realistic tool call result

RESPONSE GUIDELINES
- Tool call is valid OR has any error other than the two specified below -> Return a normal, realistic response based on the tool call and parameters.
- Tool does not exist -> "Error: Tool 'tool_name' does not exist. Available tools are: [...]"
- Undefined parameter provided -> "Error: Parameter '[parameter_name]' is not defined in the tool schema."

Keep your response simple, informative, and reasonable. Only return the tool feedback or error message, nothing else.

The tool's schema is:
{tool_schema}
The tool call parameter that awaits your feedback is:
{tool_name}: {param}
\end{promptbox}

\noindent\textbf{Modification augmentation (targeted).} Because Task Modification samples are a minority, we run a dedicated augmentation that materializes a creation-type tool call into the history and asks the LLM to generate a user modification utterance plus the corresponding modify, delete, or cancel tool call as the new expected calls.
\begin{promptbox}
You are the augmenter for a mobile-assistant tool-calling dataset. You output strict JSON only, with no explanation.

Task: given an already-completed creation-type tool call, generate the user's natural next-turn change of mind, together with the existing modify / delete / cancel tool that should be invoked.

Hard requirements:
1. target_tool must be chosen from allowed_target_tools; never invent a new tool name.
2. arguments may only use parameter names that already exist in the chosen target_tool schema; using any parameter outside the schema is a failure.
3. final_user must be natural colloquial user speech and must refer to the object that was just created, using a "just now / just set / just created" style reference.
4. final_user must also carry a change-of-mind signal, such as "change to / adjust / switch to / forgot to mention / add / cancel / no longer want".
5. [Literal consistency -- most important] every argument value in arguments must already appear verbatim in final_user or created_arguments, copied exactly, character for character:
   - the value of a new-value parameter (updated_datetime / create_event / location_new, etc.) must appear verbatim in final_user;
   - the value of an old-value parameter (activity / event / incident / occurrence / date_time / position / occasion, etc.) must appear verbatim in final_user or created_arguments;
   - never convert a relative expression in final_user (e.g. "tomorrow / next Monday / the day after tomorrow") into an absolute date; whatever final_user says, arguments copies;
   - never rewrite, expand, normalize or complete characters that are not present in final_user (for example, do not expand a short object name into a longer one, and do not turn a bare clock time into a full timestamp);
   - if final_user does not provide some new value, do not invent one: either leave that parameter out or adjust final_user so that the new value is written in.
6. For enum parameters such as type_of_range / status / ignore_holiday_state, use a value from their enum; a "just now" reference normally points to the object created in the previous round.
7. answer must be an array, and each object inside it must contain result_content and show_string, both as JSON strings.
8. ground_truth is the short summary the assistant gives the user after the tool completes; the time / place / event names appearing in it must also match final_user verbatim and must not be normalized.

Output JSON format:
{
  "final_user": "...",
  "target_tool": "...",
  "arguments": {...},
  "answer": [{"result_content": "{...}", "show_string": "{...}"}],
  "ground_truth": "..."
}
\end{promptbox}

\subsection{Self-Generated Context Manager Prompts}
\label{app:prompt_selfcm}

\noindent\textbf{CM generation prompt.} During rollout, the policy drafts a Context Manager block as the conditioning signal for the subsequent tool-call generation. The updater $\mathcal{U}$ first maintains the BeliefState $\mathbf{b}_t$ from the observable dialogue messages and tool returns (no gold intent / gold slot / gold stale label access; the same updater runs at inference). The prompt below is then assembled from $\mathbf{b}_t$ and the recent dialogue in six parts: a fixed nine-block field-semantics guide, the recent-dialogue tail together with the latest user message, the previous action, the rule-based state summary, an external hint on whether a tool is needed this turn, and a structural anchor produced by rendering the deterministic XML template of the current state. The anchor gives the model a correctly shaped skeleton to correct rather than a blank slate. The drafted tokens $\mathbf{c}^{\text{model}}_t$ are injected into the system prompt and condition the tool-call response $a_t$.
\begin{promptbox}
You are a dialogue state extractor. Based on the current dialogue state, output the Context Manager.
The Context Manager is not a reply to the user, but a compressed dialogue state for you to read
before deciding the next action.

## Sub-block structure (9 blocks, fixed order)
<Current_User>                        the user's latest utterance + intent type + whether a tool is needed this turn
<Overall_User_Task>                   overall goal + what this turn must accomplish + shelved intents
<History_Relation>                    relation to history (which slots were inherited / overridden / deleted)
<Resolved_References>                 resolved entity references (what "it" or "that one" points to)
<Active_Objects>                      still-valid tool results, keyed obj_<tool>_<index>
<Pending_Or_Interrupted_Tasks>        tasks the user raised but that remain unfinished (interrupted ones go here)
<Task_Graph_For_Current_Turn>         concrete steps for this turn: tool / arguments / depends_on / status ...
<Deferred_Tasks_After_Current_Turn>   sub-goals to do after this turn
<Tool_Decision_Guide>                 whether to call a tool + which arguments must be kept/overridden + unresolved slots

## Field semantics
- Current_User.text: the verbatim latest user message. Do not leave it empty.
- Current_User.intent_type: how this turn relates to previous ones (new_task / modify_previous_task /
  switches_topic / cancel_previous_task / ambiguous_modify_request / append_subtask).
- current_turn_requires_tool / call_tool_now / ask_user_instead: use null when unsure; do not guess false.
- Active_Objects: one entry per still-relevant tool result, each value carrying
  type / status / canonical_slots / last_tool / last_result.
- must_override_arguments and stale_arguments: when the user changes a parameter set earlier, the old
  value is NOT deleted but kept as a stale record so the agent does not regress to it.
  must_override_arguments lists the stale keys; stale_arguments gives the old value each key was
  overridden from (e.g. after the destination is overridden from Beijing to Shanghai, write
  {"destination": "Beijing"}). The NEW value goes into this turn's arguments (Task_Graph /
  Active_Objects); the old value appears only in stale_arguments.
- Task_Graph_For_Current_Turn: concrete steps this turn, each with task_id / description / tool /
  arguments / depends_on / can_parallel / status / execute_in_current_assistant_turn.
- Pending_Or_Interrupted_Tasks and Deferred_Tasks_After_Current_Turn: tasks the user raised earlier
  that are not yet finished (including interrupted ones).

## Rules
- Every JSON value must be valid JSON (double quotes, no trailing commas).
- Anything absent from the dialogue must be "unknown" / null / an empty array; do not fabricate.
- Keep the tag names, order, and JSON field names of the 9 sub-blocks unchanged; only fix or complete contents.

## Recent dialogue
{chat_tail}

## Latest user message
{last_user_message}

## Previous action
{last_action_summary}

## Rule-based state summary (may be incomplete or wrong)
{state_summary}

## Whether a tool is needed this turn (external hint)
{hint_line}

## Structural anchor — template rendering of the current state
This is the template's best guess of the current state. Correct or complete it using the field
semantics above, keeping tag names, order, and JSON field names unchanged. Empty values in the anchor
(e.g. an empty Current_User.text) do NOT mean they should stay empty; fill them in by semantics.
{template_text}

Output ONLY the content of the <Context_Manager> block: start from the first sub-block <Current_User>,
do not emit the <Context_Manager> opening tag, and end with </Context_Manager> after the last sub-block
</Tool_Decision_Guide>. Output nothing else — no explanation, no markdown fences.
<Context_Manager>
\end{promptbox}

\noindent\textbf{CM injection hint.} The drafted CM block is appended to the system prompt with a usage hint that clarifies its role and conflict priority: the CM is structural context, not a user request, and any conflict is resolved in favor of the current user message and tool schema.
\begin{promptbox}
---
# Context Manager Usage
The system provides a <Context_Manager> state block before the current assistant response.
It is not a user request, but structured context to assist your tool-call decisions.
If the Context_Manager conflicts with the current user message or tool schema, the current user message and tool schema take precedence.
Do not output the <Context_Manager> content in your final answer.
---

# Context Manager for the Current Turn
\end{promptbox}

\noindent\textbf{Intent classification (LLM fallback).} When the rule-based keyword classifier conflicts or is unconfident, an LLM classifies the user turn into one of six intent categories and extracts overridden stale slots. The JSON output is consumed by the BeliefState updater $\mathcal{U}$, which then mutates $\mathbf{b}_t$ accordingly.
\begin{promptbox}
You are an intent classifier for a multi-turn AI agent.

Given the current agent state and the user's latest message, classify the user's intent into EXACTLY ONE category:

- modified: user CHANGES a previously set parameter (e.g. "change it to 8 o'clock", "change to X")
- interrupted: user SWITCHES to a completely different task mid-flow (e.g. "wait, first help me with XX", "wait, do X instead")
- reset: user wants to START OVER / cancel (e.g. "never mind, let's not do it", "cancel")
- clarify: user ASKS for information / doesn't know what's available (e.g. "what's available?", "what can you do?")
- accumulate: user ADDS more steps to the current task (e.g. "then help me...", "based on the above...")
- none: user is continuing naturally, no intent change

Output ONLY a JSON object:
{"intent":"<one of the six>","confidence":<0.0-1.0>,"reason":"<1 sentence>",
 "extracted_slots":{"key":"value"},"new_goal":"<if interrupted/reset>",
 "stale_slot_keys":["keys that were overridden"]}
\end{promptbox}

\noindent\textbf{CM render XML template.} The target CM format represents the BeliefState using nine XML sub-blocks, each a JSON object. During training, a deterministic template renders a reference CM from the corresponding BeliefState field; it supplies the structural anchor in the generation prompt above and is not itself a loss target, and the model is trained to produce tokens matching this format from $(\mathbf{b}_t, h_t)$. The skeleton is shown below; field semantics follow the BeliefState definition in Section~4.2.
\begin{promptbox}
<Context_Manager>
  <Current_User>{"text": "<user's latest utterance>",
    "intent_type": "new_task | modify_previous_task | switches_topic |
      cancel_previous_task | ambiguous_modify_request | append_subtask",
    "current_turn_requires_tool": <bool|null>}</Current_User>
  <Overall_User_Task>{"full_intent": "<current_goal>",
    "current_turn_boundary": "<what this turn does>",
    "deferred_intent_summary": "<or empty>"}</Overall_User_Task>
  <History_Relation>{"relation_to_history": "none|modify|switch|append",
    "target_object_ids": [<ids>], "inherit_slots": [<active>],
    "override_slots": [<stale>], "delete_slots": [<deleted>],
    "reason": "<intent>"}</History_Relation>
  <Resolved_References>{}</Resolved_References>
  <Active_Objects>{"obj_<tool>_<turn>": {"type": "tool_result",
    "status": "active", "canonical_slots": {...}, "last_tool": "<name>",
    "last_result": "<summary>"}}</Active_Objects>
  <Pending_Or_Interrupted_Tasks>[{"task_id": "P1",
    "description": "<pending>", "status": "blocked"}]
    </Pending_Or_Interrupted_Tasks>
  <Task_Graph_For_Current_Turn>[{"task_id": "T1", "description": "<goal>",
    "tool": "<name>", "arguments": <active_slots>, "depends_on": [],
    "can_parallel": false, "status": "ready",
    "execute_in_current_assistant_turn": true}]</Task_Graph_For_Current_Turn>
  <Deferred_Tasks_After_Current_Turn>[{"task_id": "D1",
    "description": "<deferred_goal>", "status": "deferred",
    "reason": "<reason>"}]</Deferred_Tasks_After_Current_Turn>
  <Tool_Decision_Guide>{"call_tool_now": <bool|null>,
    "ask_user_instead": <bool|null>, "expected_tool_count": <int>,
    "must_call_tools": [<names>], "must_not_call_tools": [<names>],
    "must_preserve_arguments": [<keys>],
    "must_override_arguments": [<stale_keys>],
    "stale_arguments": {"<key>": "<old_value>"},
    "unresolved_slots": [<questions>],
    "common_error_warnings": [<warnings>]}
    </Tool_Decision_Guide>
</Context_Manager>
\end{promptbox}

\subsection{Reward and Evaluation Judge Prompts}
\label{app:prompt_judge}

The hierarchical reward uses LLM-as-a-judge for the two semantically ambiguous signals (ACR and ISSR); SCRR and DTCR are rule-based in the reward. The \emph{evaluation} suite uses a more fine-grained set of judge prompts (8-level ACR, 7-level ISSR, plus SCRR and DTCR) for diagnosis. We list the reward-side prompts and one evaluation-side prompt (ISSR) for completeness; the others follow the same structure with extended rubrics.

\noindent\textbf{ACR judge (reward-side, active clarification).} Scores whether the agent's clarification response matches the ground-truth intent in $[-1,1]$; a positive score is scaled by $R_{\text{ACR}}{=}1.0$ into $R_{\text{action}}$.
\begin{promptbox}
You are an expert evaluator for AI agent clarification quality.
Evaluate whether the agent's response matches the ground-truth intent.

## Scoring Rubric (score in [-1.0, 1.0])
### 1.0 — Perfect: exact semantic match, covers ALL missing params
### 0.7-0.9 — Near Perfect: core intent aligned, minor wording differences
### 0.4-0.6 — Good: main intent aligned, covers most params
### 0.0-0.3 — Weak: partial alignment
### -0.1 to -0.5 — Mostly Irrelevant
### -0.6 to -1.0 — Completely Wrong / hallucinated

Return ONLY: {"score": <float between -1.0 and 1.0>, "reasoning": "<1 sentence>"}
\end{promptbox}

\noindent\textbf{ISSR judge (reward-side, intent switch).} Scores whether the agent successfully switched from the old task to the new one in $[-1,1]$; scores $>0.3$ are weighted by $R_{\text{ISSR}}{=}2.0\,s_{\text{LLM}}$ and added into $R_{\text{outcome}}$ (score-weighted, not a fixed pivot bonus).
\begin{promptbox}
You are an expert evaluator for multi-turn agent trajectory analysis.
Determine whether the agent successfully SWITCHED from old task to new task.

## Scoring Rubric (score in [-1.0, 1.0])
### 1.0 — Perfect Switch: fully abandoned old, correctly executed new
### 0.5-0.9 — Good Switch: mostly switched, minor residue
### 0.0-0.4 — Weak / Unclear
### -0.1 to -0.5 — Mostly Stuck on Old
### -0.6 to -1.0 — Completely Ignored Interruption

Return ONLY: {"score": <float between -1.0 and 1.0>, "reasoning": "<1 sentence>"}
\end{promptbox}

\noindent\textbf{ISSR judge (evaluation-side, fine-grained).} The evaluation suite uses an extended 7-level rubric covering abandonment, execution, transition and completeness, and emits flags for interruption detection and switch success to support diagnostic aggregation.
\begin{promptbox}
You are an expert evaluator for multi-turn AI agent trajectory analysis.
Your task: determine whether the agent successfully SWITCHED to the new user intent after the user interrupted or shifted the topic in the middle of a previous task.

## Context
- The agent was executing OLD TASK A.
- The user suddenly interrupted with NEW TASK B (or an unrelated question).
- The agent should ABANDON Task A and execute Task B.
- Failure mode: the agent gets "stuck" on Task A and either ignores Task B or tries to merge them incorrectly.

## Evaluation Dimensions
1. Abandonment of Old Task — Did the agent STOP executing Task A?
2. Execution of New Task — Did the agent correctly execute Task B?
3. Transition Quality — Did the agent acknowledge the switch gracefully?
4. Completeness — For Task B, did the agent do everything required?

## Scoring Rubric (score in [-1.0, 1.0])
### 1.0 — Perfect Switch: completely stopped A, fully executed B.
### 0.7~0.9 — Good Switch, Minor Residue: switched to B, minor A reference.
### 0.4~0.6 — Partial Switch: attempted B but incomplete, noticeable A residue.
### 0.0~0.3 — Weak / Unclear: acknowledged but barely acted on B.
### -0.1~-0.3 — Mostly Stuck on Old: continued A, lip service to B.
### -0.4~-0.6 — Confused Mix: merged A and B incoherently, called tools for both.
### -0.7~-1.0 — Completely Ignored Interruption: continued A as if nothing happened.

## Output Format
Return ONLY a JSON object:
{"score": <float>, "is_interruption": true/false, "switch_success": true/false, "reason": "1-2 sentence justification"}

## Input Data
Previous-turn assistant tool calls: {prev_tool_calls}
Previous-turn assistant response: {prev_response}
Current-turn user input: {curr_user}
Current-turn model tool calls: {pred_tool_calls}
Current-turn model response: {pred_response}
Note: switch_success is only meaningful when is_interruption=true.
\end{promptbox}

\bibliography{aaai2027}

\endgroup

\end{document}